\documentclass[10pt,twocolumn,letterpaper]{article}

\usepackage{cvpr}              
\usepackage[accsupp]{axessibility}

\newcommand{\TODO}[1]{\textbf{\color{red}[TODO: #1]}}
\renewcommand{\TODO}[1]{}

\usepackage{microtype}

\renewcommand{\paragraph}[1]{\vspace{.5em}\noindent\textbf{#1.}}

\usepackage{soul}
\setuldepth{foobar}
\definecolor{dartmouthgreen}{rgb}{0.05, 0.5, 0.06}
\definecolor{darkviolet}{rgb}{0.58, 0.0, 0.83}
\usepackage{tabularx}
\usepackage{tcolorbox}

\definecolor{cvprblue}{rgb}{0.21,0.49,0.74}
\usepackage[pagebackref,breaklinks,colorlinks,allcolors=cvprblue]{hyperref}

\def\paperID{44105} 
\def\confName{CVPR}
\def\confYear{2026}

\title{LVLM-Aided Alignment of Task-Specific Vision Models}

\author{
Alexander Koebler$^{1}$\thanks{Corresponding author: alexander.koebler@gmx.de} \quad
Lukas Kuhn$^{1,3,4}$ \quad
Ingo Thon$^{2}$ \quad
Florian Buettner$^{1,3,4}$\\[0.5em]
$^{1}$Goethe University Frankfurt \quad
$^{2}$Siemens AG \\
$^{3}$German Cancer Research Center (DKFZ) \quad
$^{4}$German Cancer Consortium (DKTK)\\
}

\begin{document}
\maketitle
\begin{abstract}
    In high-stakes domains, small task-specific vision models are crucial due to their low computational requirements and the availability of numerous methods to explain their results. However, these explanations often reveal that the models do not align well with human domain knowledge, relying instead on spurious correlations. This might result in brittle behavior once deployed in the real-world.
    To address this issue, we introduce a novel and efficient method for aligning small task-specific vision models with human domain knowledge by leveraging the generalization capabilities of a Large Vision Language Model (LVLM). Our LVLM-Aided Visual Alignment (LVLM-VA) method provides a bidirectional interface that translates model behavior into natural language and maps human class-level specifications to image-level critiques, enabling effective interaction between domain experts and the model.
    Our method demonstrates substantial improvement in aligning model behavior with human specifications, as validated on both synthetic and real-world datasets. We show that it effectively reduces the model’s dependence on spurious features and on group-specific biases, without requiring fine-grained feedback.
\end{abstract}

\section{Introduction}
\begin{figure}
    \includegraphics[width=1.0\columnwidth]{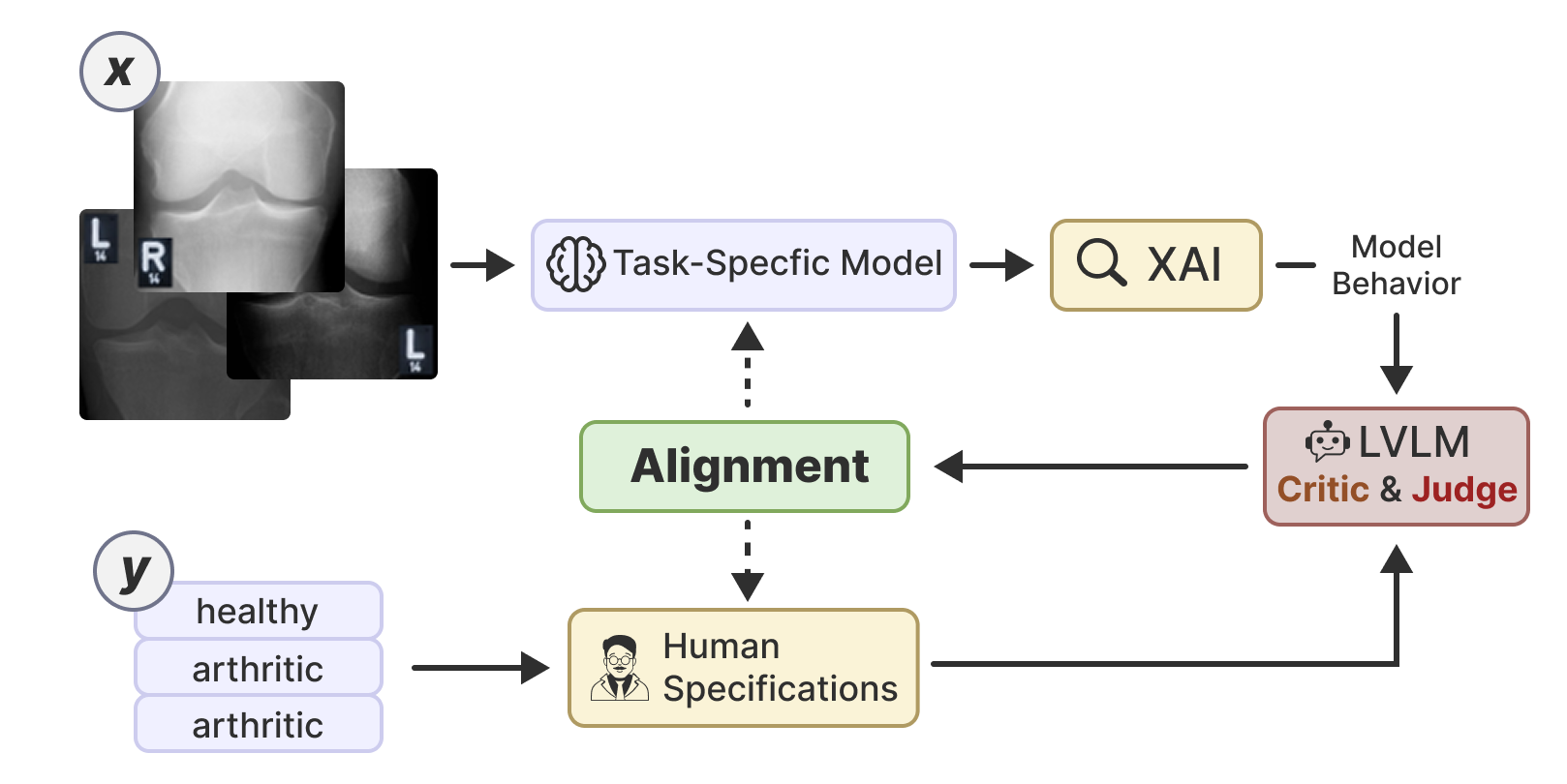}
    \caption{LVLM-Aided Visual Alignment (LVLM-VA) of a small task-specific vision model steered by human domain knowledge, using Explainable AI (XAI) in conjunction with a Large Vision Language Model (LVLM) Critic \& Judge pair. The domain knowledge is induced into the system via human specifications on a class level supporting the LVLM to identify relevant core features within the input images and detect spurious shortcuts based on the model explanations. The Critic \& Judge assessment is used to correct the original model in an alignment step but can also be used to provide feedback to the human expert.}
    \label{fig:fig1}
\end{figure}
In an era of increasingly large general-purpose models being able to interpret and translate visual inputs in natural language, reliable small task-specific vision models for narrow classification tasks are still of vital importance. This is especially true in many high-stakes domains where interpretability and trustworthiness demands are rigorous. Examples include the medical and manufacturing domains, where misclassifications can have severe downstream impact, requiring high robustness and explainability. For these non-functional requirements, current Large Vision Language Models (LVLMs) fall short \citep{Guan_2024_CVPR, yang2021causal}.\\
However, ensuring the continued reliability of small task-specific models and making their predictions interpretable to subject-matter experts remain challenging \citep{decker2023thousand}. Spurious correlations in relatively small training datasets for narrow tasks can cause a model to learn shortcuts that yield good performance on the training distribution but result in brittle behavior when the model is deployed in the real world \cite{lapuschkin_clever_hans, rueckel_conf_med}.
One way to tackle this issue and increase the reliability of models is to explicitly incorporate human domain knowledge into the model training pipeline \citep{von2021informed} and by this align the model with how a human would solve the task. While Explainable AI (XAI) techniques can be utilized to identify the learned shortcuts and make targeted corrections, interpreting the explanations generated by widely used XAI methods is often very difficult for domain experts. Furthermore, current methods rely on instance-wise feedback on the model's explanations \cite{schramowski2020making, ross2017right}, which is too time-consuming for experts who are often highly specialized professionals, such as medical doctors.\\
However, the bidirectional process of aligning an ML model, consisting of providing feedback and interpreting the model's reasoning, is important not only for incorporating human knowledge and values but also for increasing user trust \citep{shen2024towards}.\\\\
In this work, we introduce a synergistic approach that leverages recent advances in the capabilities of LVLMs to align small, task-specific vision models with human domain knowledge. The LVLM acts as a bidirectional translator. First, it translates explanations of the current model’s behavior from image space into natural language, highlighting spurious correlations. Second, it translates human domain knowledge about the vision task, expressed in natural language, into instance-wise critiques in image space. Thus, the LVLM provides domain experts with a more intuitive interface through which they can actively steer the model and critically evaluate its reasoning.
To achieve this, we make the following contributions:
\begin{itemize}
    \item We propose LVLM-Aided Visual Alignment (LVLM-VA) as a novel approach allowing for automated instance-wise correction from class-level human specifications to efficiently align a neural network with human domain knowledge, reducing its reliance on spurious correlations.
    \item We introduce Positive Predictive Effect Probabilistic Segmentation via Weighted Gaussian Mixtures (PPEPS-WGM) to facilitate an LVLM to translate model behavior into natural language to detect spurious features. 
    \item We demonstrate on different synthetic and real-world scenarios that our approach can effectively reduce the reliance of vision models on shortcuts by aligning them with human domain knowledge, without requiring any fine-grained feedback.
\end{itemize}

\section{Related Work}
Previous works have addressed the challenge of debugging models relying on spurious correlations by fine-tuning the model with human critique based on explanations of the current model behavior \citep{teso_xil, ross2017right}. Thereby, these methods improve the alignment of the model with human reasoning. However, they often require extensive fine-grained feedback for each image \citep{schramowski2020making}. Furthermore, explanations of the current model behavior and feedback on potential errors must be provided directly in the image space \citep{ross2017right}. This results in an inefficient interaction with the model.
\citet{stammerlearning} introduce a method to allow a Vision Language Model (VLM) to internally critique its own explanations independent of external human input increasing the model's performance but not explicitly aiming for aligning the model with human domain knowledge. Furthermore, the method does not translate to general small task-specific vision models for broad classification tasks. 
The authors in \citet{zheng2024learning} utilize a general-purpose captioning model to extract textual concepts without human steering from images and define a spuriousness score for each concept based on the accuracy of the classifier with and without that concept. However, the proposed captioning models might be incapable of identifying concepts for settings not explicitly included in the training data, e.g., medical or industrial images, and are limited to relatively discrete, co-occurring concepts, making it unsuitable for spurious features that manifest as subtle, continuous variations such as slight color differences.
\citet{gu2024anomalygpt} introduce an approach to use an LVLM to provide explanations of the model's decision in natural language. However, they do not consider the natural language interface to inject human feedback back into the model. In contemporary work, \citet{kuhn2025efficient} proposed a highly specialized VLM-based method for mitigating shortcuts in vision transformers.\\
Aside from the previously mentioned approaches, non-human-centred methods focus on mitigating shortcuts without describing them directly in image space. These methods instead address the issue by balancing the model's performance between groups categorized by the class label and the presence or absence of spurious features. These approaches do not require instance-wise feedback about the location of spurious features; however, they do need additional annotations about the presence of spurious features per image. \citet{kirichenkolast} propose Deep Feature Reweighting (DFR), in which they only retrain the final layer of the vision model on a small, balanced dataset. This is based on the assumption that core features are often already learned during the initial training phase and simply need to be reweighted to improve performance on the test set. \citet{idrissi2022simple} demonstrate that straightforward data balancing techniques, such as subsampling or reweighting based on group frequencies, can deliver competitive worst group accuracy without the need for sophisticated training procedures. Lastly, \citet{liu2021just} introduce Just Train Twice (JTT), whereby an initially on a few epochs trained model identifies challenging examples, and a second model is then trained on a reweighted dataset that up-samples these examples, aiming to reduce reliance on spurious correlations. These non-human-centric methods solely target equal performance across groups, regardless of the features used to achieve it. More precisely, those methods do not aim to directly align the model with human domain knowledge and thus lack explainability.

\section{Problem Setting}
Assume we have a vision model $f: \mathcal{X} \rightarrow \mathcal{Y}$ trained on a labeled training dataset $D_s = {(x_s, y_s)}^{n_s}_{i=1}$ with input images $x_s \in \mathbb{R}^{C \times H \times W}$ and labels $y_s \in \{1,...,K\}$. For every class there exists a human specification $\mathcal{V}_k$ elaborating on important features to identify the specific class $k$. Further, explanation function $\Phi: \mathcal{X} \times \mathcal{Y} \times f \rightarrow \mathbb{R}^{H\times W}$ generates explanation maps in the original image space indicating which regions the model $f$ considers important for predicting the output $y$ given the input $x$. However, the trained model $f$ might focus on areas identified by the explanations $\Phi(x,y,f)$ which do not match the description $\mathcal{V}_k$ indicating insufficient alignment of the model with domain knowledge and the reliance on spurious features. This results in reduced performance when the model is applied to test samples $D_t = {(x_t, y_t)}^{n_t}_{i=1}$ not subject to spurious correlations.

\section{LVLM-Aided Visual Alignment (LVLM-VA)}
We introduce our LVLM-Aided Visual Alignment (LVLM-VA) method as a two step approach to reduce the reliance of any vision classification model on spurious features.

\subsection{Detecting Spurious Correlations}
\begin{figure}
\centering
\includegraphics[width=1.0\columnwidth]{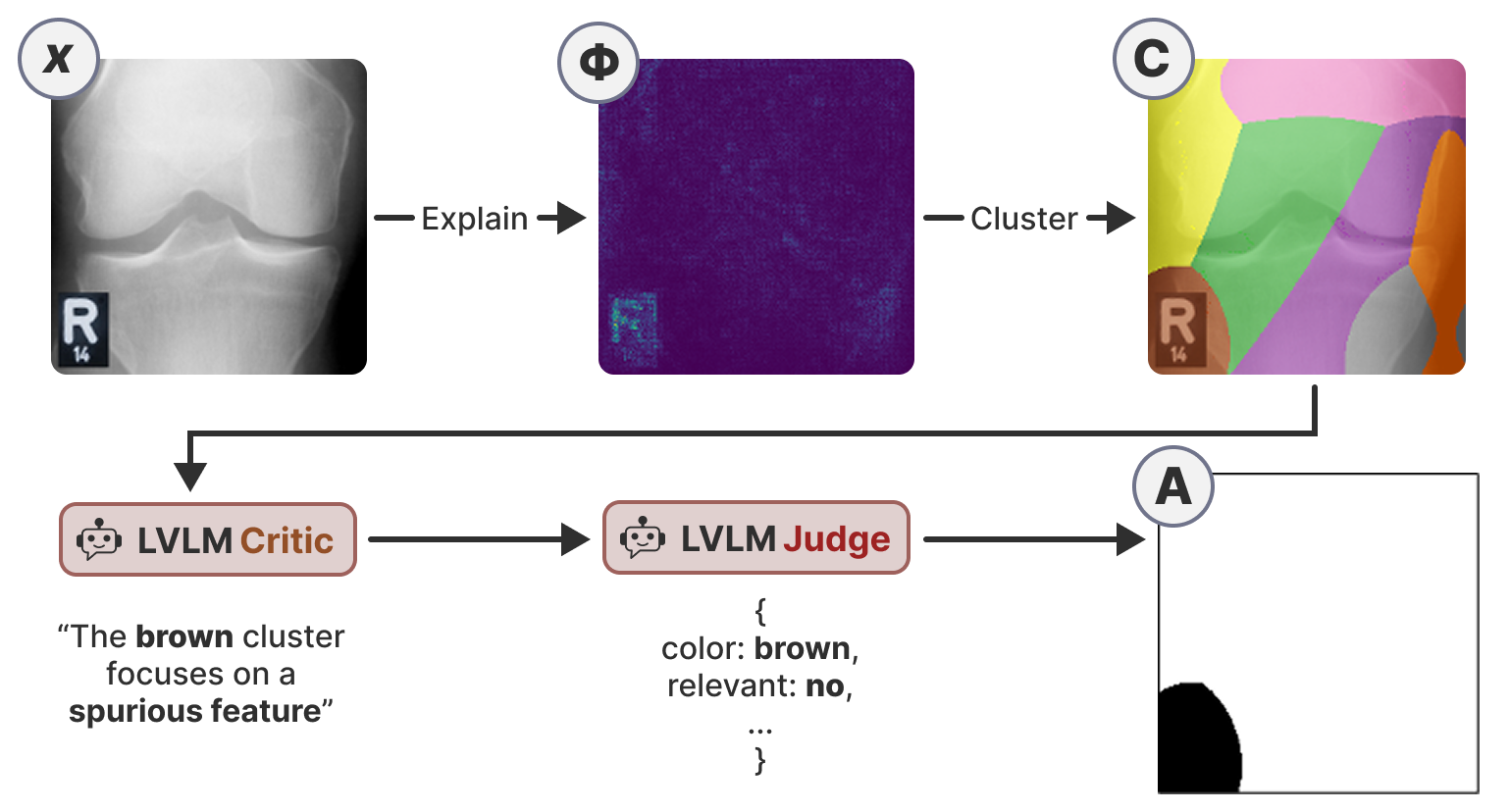}
\caption{Correction mask generation process by the Critic \& Judge pair for a vision model trained on a knee radiograph dataset. The image shows a hospital tag in the bottom left, which the model learned as a shortcut to classify the condition of the knee. All images are of size $224\times 224$.}
\label{fig:exp}
\end{figure}

In the initial steps of our LVLM-VA approach (\cref{fig:exp}), a combination of XAI and an LVLM is used to generate an instance-wise correction signal aligning class-level human specifications and instance-level model explanations.\\
\textbf{Sampling Strategy:}
While our procedure for detecting spurious features and aligning the original model can, in principle, be applied to the full training set, doing so would cause an unnecessary computational overhead on large datasets due to the reliance on a LVLM. To address this, we preferably generate the steering signal only for those samples on which the model relies on shortcuts.
We propose an unsupervised sampling strategy based on the model’s output entropy, motivated by the assumption that shortcuts are easier to learn than robust core features \cite{hermannfoundations}. As a result, the model tends to exhibit lower output entropy on shortcut dependent training examples. Consequently, in subsequent steps, our alignment dataset $D_{\text{align}}$ is defined as the $N$ training samples with the lowest output entropy under the original model $f$. In contrast to other shortcut-mitigation approaches \cite{kirichenkolast, idrissi2022simple}, our method does not require additional group labels to identify samples affected by spurious features.\\
\textbf{Positive Predictive Effect Probabilistic Segmentation via Weighted Gaussian Mixtures (PPEPS-WGM):}

Based on the alignment set \(D_{\text{align}}\), we use a large vision language model (LVLM) to identify potential spurious features. Following \citet{yang2023set}, we introduce a pre-segmentation step on the alignment images. The authors in  \cite{yang2023set} employ a Segment Anything Model (SAM) \cite{kirillov2023segany} to partition images based on visual content, showing that such pre-segmentation improves LVLM's capability of spatial location. However, we seek segmentation to support the detection of spurious features. Thus, we generate explanation maps \(\Phi(x,y,f)\) in image space on the alignment dataset \(D_{\text{align}}\) as a proxy for the current decision process of the vision model \(f\). Following, we introduce \emph{Positive Predictive Effect Probabilistic Segmentation via Weighted Gaussian Mixtures (PPEPS-WGM)}, which performs model-centric segmentation by fitting a weighted Gaussian mixture whose components cluster regions according to their positive predictive effect shown in \(\Phi(x,y,f)\). This steers the pre-segmentation and the LVLM's subsequent spatial allocation toward regions that most influence \(f\)'s predictions rather than object boundaries, enabling more targeted identification of spurious features, i.e., regions of high positive attribution.

\textbf{Setting:}
Let $f$ be a trained model and let $b(x)$ denote the \emph{additive} quantity we explain, i.e., the output logit of the target class. As an explanation method, we use DeepLIFTSHAP \cite{DeepShap} because of the convenient theoretical properties of Shapley values.
More precisely, given an input image \(x\in\mathbb{R}^{H\times W}\) with \(M=H\cdot W\) pixels indexed by \(i\in\{1,\dots,M\}\), Shapley values produce per-pixel attributions \(\{\Phi_i(x)\}_{i=1}^M\) that satisfy \emph{local accuracy}:
\begin{align*}
\label{eq:local-accuracy}
\sum_{i=1}^{M} \Phi_i(x) \;=\; b(x) - \mathbb{E}[b(X)].
\end{align*}
Thus, $\{\Phi_i\}$ form a finite signed measure over pixels, where the attribution of a region equals the sum of its constituents and the total effect mass is conserved.
In the following steps we focus on positive contributions as we consider spurious correlations as signals that falsely increase the score of the given prediction.
We define the positive part of the model attribution and its total mass as:
\begin{align*}
\Phi_i^{+}(x) \;=\; \max\{\Phi_i(x),\,0\}, 
\qquad 
Z^{+}(x) \;=\; \sum_{i=1}^{M} \Phi_i^{+}(x).
\end{align*}
We subsequently normalize $\Phi_i^{+}(x)$ to obtain a discrete probability mass function (PMF) over pixels
\begin{align*}
p_i(x) \;=\; \frac{\Phi_i^{+}(x)}{Z^{+}(x)}, 
\qquad 
\sum_{i=1}^M p_i(x)=1.
\end{align*}
With that \(p_i(x)\) is the probability that a randomly sampled unit of \emph{positive predictive effect} (positive additive change in \(b\)) lies at pixel \(i\).\\

\textbf{Weighted Gaussian Mixture on the Positive-Effect Distribution:}
Let \(z_i\in\mathbb{R}^d\) denote per-pixel features used for clustering. 
As we would like to segment where in the image the positive effect mass lives, not what the underlying content is, we instantiate \(z_i\) as \emph{normalized spatial coordinates} on the equidistant image lattice,
\[
z_i=\big((u_i+\tfrac12)/H,\; (v_i+\tfrac12)/W\big)\in[0,1]^2,
\]
with \((u_i,v_i)\in\{0,\ldots,H\!-\!1\}\times\{0,\ldots,W\!-\!1\}\) the pixel indices. %
We fit a $J$-component Gaussian mixture to the discrete distribution \(p\) by weighted maximum likelihood:\\
\begin{align*}
\label{eq:wgmm-objective}
\max_{\Theta}\; &\mathcal{L}(\Theta)
\;=\;
\sum_{i=1}^{M} w_i \,
\log\!\Bigg(\sum_{j=1}^{J} \pi_j \, \mathcal{N}\!\big(z_i\mid \mu_j,\Sigma_j\big)\Bigg),\\
&w_i \;=\; M \cdot p_i(x),
\end{align*}
where \(\Theta=\{(\pi_j,\mu_j,\Sigma_j)\}_{j=1}^J\) are the mixture parameters with \(\pi_j\ge0\) and \(\sum_j \pi_j=1\).
We define the responsibilities
\begin{align*}
r_{ij} \;=\; 
\frac{\pi_j \,\mathcal{N}(z_i\mid \mu_j,\Sigma_j)}{\sum_{\ell=1}^J \pi_\ell\,\mathcal{N}(z_i\mid \mu_\ell,\Sigma_\ell)}.
\end{align*}
The M-step sufficient statistics are formed with effective weights $w_i r_{ij}$ for pixel $i$ and component $j$.
The mixture-weight update equals the positive-effect share captured by component \(j\):
\begin{align*}
\pi_j^{\mathrm{new}}
\;=\;
\frac{\sum_{i=1}^M w_i r_{ij}}{\sum_{i=1}^M w_i}
\;=\;
\sum_{i=1}^M p_i(x) \, r_{ij}
\;=\; S_j,
\end{align*}
where \(S_j\in[0,1]\) and \(\sum_j S_j=1\).
Hence \(\pi_j\) at optimum is interpretable as the \emph{fraction of total positive effect} explained by component \(j\).

Lastly, we derive a segmentation by assigning each pixel to its most probable component:
\begin{align*}
c_i &\;=\; \arg\max_{j\in\{1,\dots,J\}} r_{ij},
\qquad
C \in \{1,\dots,J\}^{H\times W}\ \\
&\text{given after reshaping } \{c_i\}_{i=1}^M.
\end{align*}
\textbf{Generating LVLM-based Alignment Verdicts:} Following the clustering step, the segmented explanation map $C$, together with the original image $x$ and the ground truth label $y$, are provided to the LVLM-based Critic $g$. Further, a third image showing the original image overlapped with the segmentation is provided to the Critic in order to support the correct localization of the segments.
To facilitate $g$ in detecting if the vision model $f$ relies on spurious features, it is instructed to utilize a chain-of-thought process \cite{wei2022chain}. The introduced prompt guides the model to (1) examine the original image, (2) identify which regions belong to the ground truth class $y$, (3) determine for each segmented cluster which parts of the original image are included, (4) combine both insights, and (5) describe if a cluster covers a relevant region, and (6) lastly provide a verdict whether a cluster is relevant based on the previous insights. To further allow to steer the LVLM in this process and emphasize what important concepts define a particular class, class-specific prompts include human-defined descriptions $\mathcal{V}_k$ about how to accurately recognize the class $k$. As these descriptions allow scaling class-level human feedback to instance-wise critique, they drastically decrease human effort for aligning the model.
To seamlessly integrate the LVLM assessment into an automatic training pipeline and reduce the overall complexity of the task, we employ an LLM Judge $h$ (which may be instantiated as the same model as $g$) that maps the free-form output of $g$ for $x$ to a final binary verdict $R$. It determines whether each cluster corresponds to a spurious feature, yielding a single binary verdict $R_{j}$ for every cluster $j$ in $C$. A class-agnostic prompt for $h$ further steers this verdict by providing example pairs of Critic assessments and their associated binary human judgments. Evaluating prototypical outputs of $g$ and specifying aligned verdicts establishes another option for bidirectional feedback, giving the human expert a mechanism to influence the final decision. In addition to aligning the outcome with human knowledge, prior work shows that such few-shot exemplars can substantially increase LLM performance on specified tasks \cite{brown2020language}.

\subsection{Visual Alignment using LVLM Verdicts}
Different previous works have focused on correcting model explanations \cite{ross2017right, slany2022caipi} by aligning them with fine-grained human feedback \citep{schramowski2020making} in the form of instance-wise corrections in image space. In our LVLM-VA approach, we utilize the Right for the Right Reasons (RRR) loss function introduced by \citet{ross2017right} for the alignment:
\begin{align*} L(\theta, X, &y, A) = \textcolor{blue}{\sum_{n=1}^{N} \sum_{k=1}^{K} -y_{nk} \log(\hat{y}_{nk})} + \\
&\textcolor{dartmouthgreen}{\lambda \sum_{n=1}^{N} \sum_{i=1}^{M} \left(A_{ni}\frac{\partial}{\partial x_{ni}} \sum_{k=1}^{K} \log(\hat{y}_{nk}) \right)^{2}} + \textcolor{darkviolet}{\gamma \sum_{i} \theta_i^2} \end{align*}
Here, $N$ is the number of used alignment samples, $K$ refers to the number of classes, and $M$ is the dimensionality of the input $x$.
The first term \textcolor{blue}{"right answers"} corresponds to the cross-entropy loss, optimizing the model to make correct classification predictions. The second term \textcolor{dartmouthgreen}{"right reasons"} ensures that the model's decisions are based on relevant features by reducing the gradient in regions deemed irrelevant by experts via a binary mask $A$, steering the model to focus on important features and avoid spurious correlations. Additionally, an optional term \textcolor{darkviolet}{"regularization"} on the model parameters $\theta$ can be added to prevent overfitting.\\
We automatically transfer the binary verdicts generated via the Critic \& Judge pair into the correction maps $A$:
\begin{equation*}
    A = \sum_{j=1}^{J} R_{j} \cdot \mathbf{1}\left[C = j\right] \; ,
\end{equation*}
where the cluster verdict $R_{j}$ is applied to the corresponding cluster $j$ in the segmented explanation map $C$ such that $A$ only features clusters considered to be spurious. By this, we render the previously required tedious per-instance interaction to generate the expert maps obsolete.
For the fine-tuning using the RRR loss, the alignment samples $x_a$ are mixed with the training samples $x_s$ in each batch of size $I$ with a ratio of $\frac{I_{x_a}}{I_{x_s}}$. An epoch for $N_{Train}$ training samples consists of $\frac{N_{Train}}{I_{x_s}}$ train iterations, in the case that this is greater than $\frac{N}{I_{x_a}}$, the alignment samples are over-sampled. With this procedure, we aim to avoid catastrophic forgetting of previously learned core features. In summary, fine-tuning the original model $f$ using the RRR loss with instance-wise masks generated automatically based on human specifications allows our method to significantly reduce manual effort whilst still allowing human steering of the vision model.
\section{Experiments}
We evaluate our approach using three different datasets and two shortcut learning settings. In the first multi-class classification setting, artificial spurious decoys occur throughout the entire training set and their appearance correlates with the classes, but they are absent in the test set, which leads to low test performance. This setting may reflect a systematic bias in the data generation process, such as different camera settings when the model is trained and deployed. In the second setting, we evaluate two real-world binary classification tasks where the spurious features remain similar in both the training and test sets, but their frequency of occurrence in the training set depends on the class, whereas they are equally distributed when the model is deployed. This could be attributed to a shift between the training and test distributions. Within these settings, we benchmark our approach against constrained optimization with instance-level human feedback \citep{ross2017right} and shortcut mitigation strategies that aim to reduce differences between group accuracies \cite{idrissi2022simple, kirichenkolast, liu2021just}. We use DeepLiftSHAP \citep{DeepShap} to generate the explanation maps $\Phi(x,y,f)$ across all experiments. For the Critic and Judge, we use a GPT-4o model. The remainder of this section first presents the multi-class setting with artificial decoys before moving on to the real-world medical setting. More details are documented in the Appendix and code is provided at: \url{https://github.com/alexanderkoebler/LVLM-VA}.
\subsection{Mitigating Systematic Decoys Across the Entire Training Set} 
\begin{figure}
    \centering
    \includegraphics[width=1.0\columnwidth]{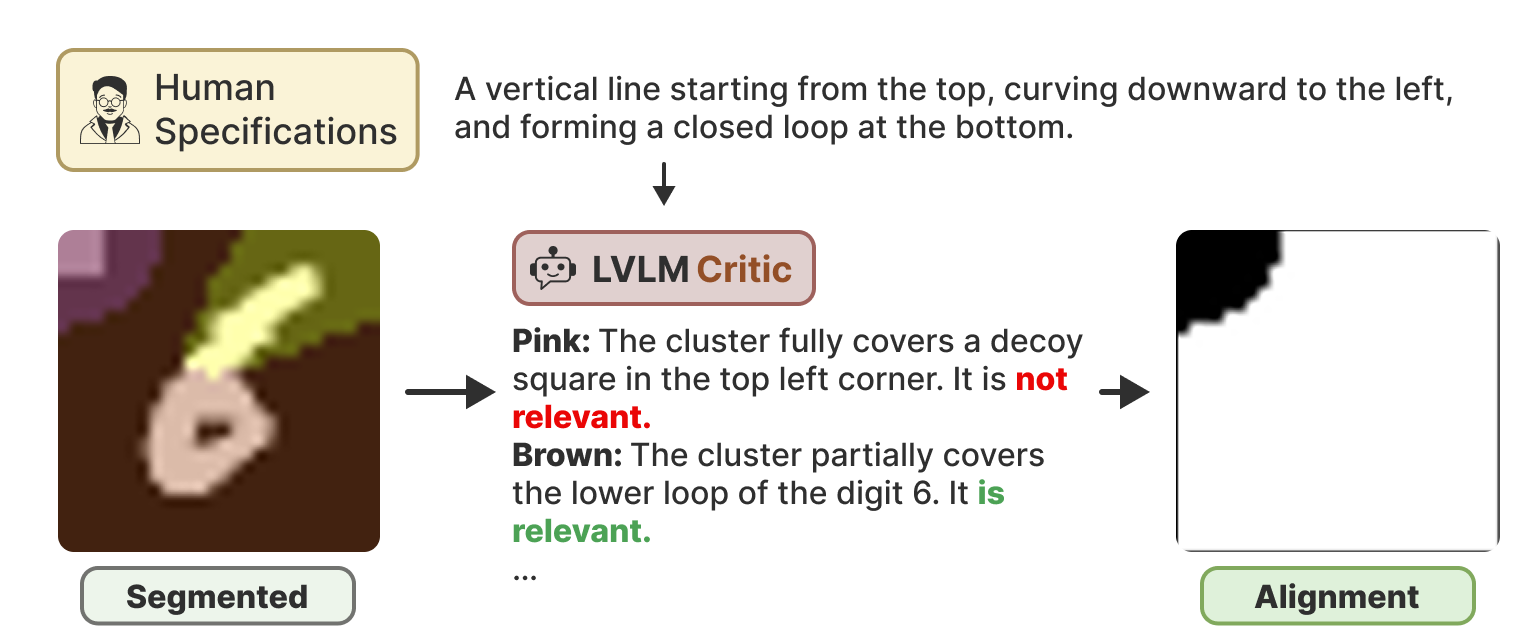}
    \caption{Intermediate results for aligning an MLP model for classifying DecoyMNIST. Based on the input of the original image, the segmentation map and the class level description, the LVLM-Critic correctly identifies that the top left corner includes the spurious decoy. The LLM-Judge assigns the right binary label which is subsequently transferred into the correction mask where black refers to 'not relevant'.}
    \label{fig:MNIST_Plots}
\end{figure}

In the first scenario, we evaluate our approach using an artificial decoy dataset based on digit classification \cite{lecun-98} with added artificial decoys. This dataset is frequently used in the literature to study model debugging \citep{ross2017right, bekkemoen2021correcting}.\\
\textbf{Experimental Setting:} Each image in the dataset contains a grey patch in a random corner. While the shade of grey for the samples in the training set depends on the digit $k$ ($255 - 25\cdot k$), it is chosen randomly in the test set. As a result, these patches represent simple shortcut candidates for the model across all classes in the training set but act as harmful confounders in the test set. For this experiment we train a two-layer Multi-Layer-Perceptron (MLP) of width $256$.
For the alignment set we use $N = 256$ samples $x_a$ from the training set using the previously described sampling strategy. The number of clusters $J$ for PPEPS-WGM is set to three which, as shown in \cref{fig:MNIST_Plots}, is sufficient for allowing to clearly identify the spurious features.
\begin{figure}
    \centering
    \includegraphics[width=0.8\columnwidth]{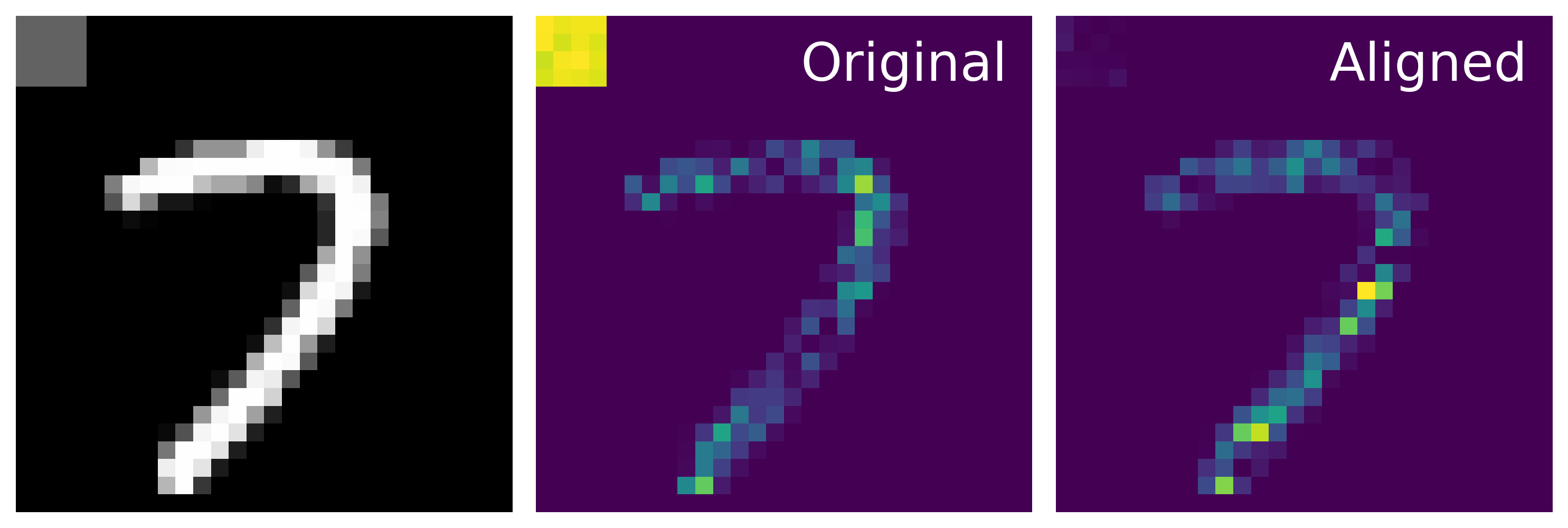}
    \caption{The explanations generated for a test example for an MLP model trained on DecoyMNIST before and after the alignment step. The original model clearly focuses on the spurious decoy in the upper left corner whereas the attribution of the aligned model is almost fully distributed across the actual digit.}
    \label{fig:exp_MNIST}
\end{figure}

\textbf{Results:} Subsequently, we first qualitatively evaluate the effect of the alignment step on the model's reliance on spurious decoys. The explanations of the MLP model in \cref{fig:exp_MNIST} show a complete reassignment of the model's attention to the actual digit. This indicates that the model is significantly less affected by the spurious features introduced during training.


\begin{figure}[h]
    \centering
    \includegraphics[width=1\columnwidth]{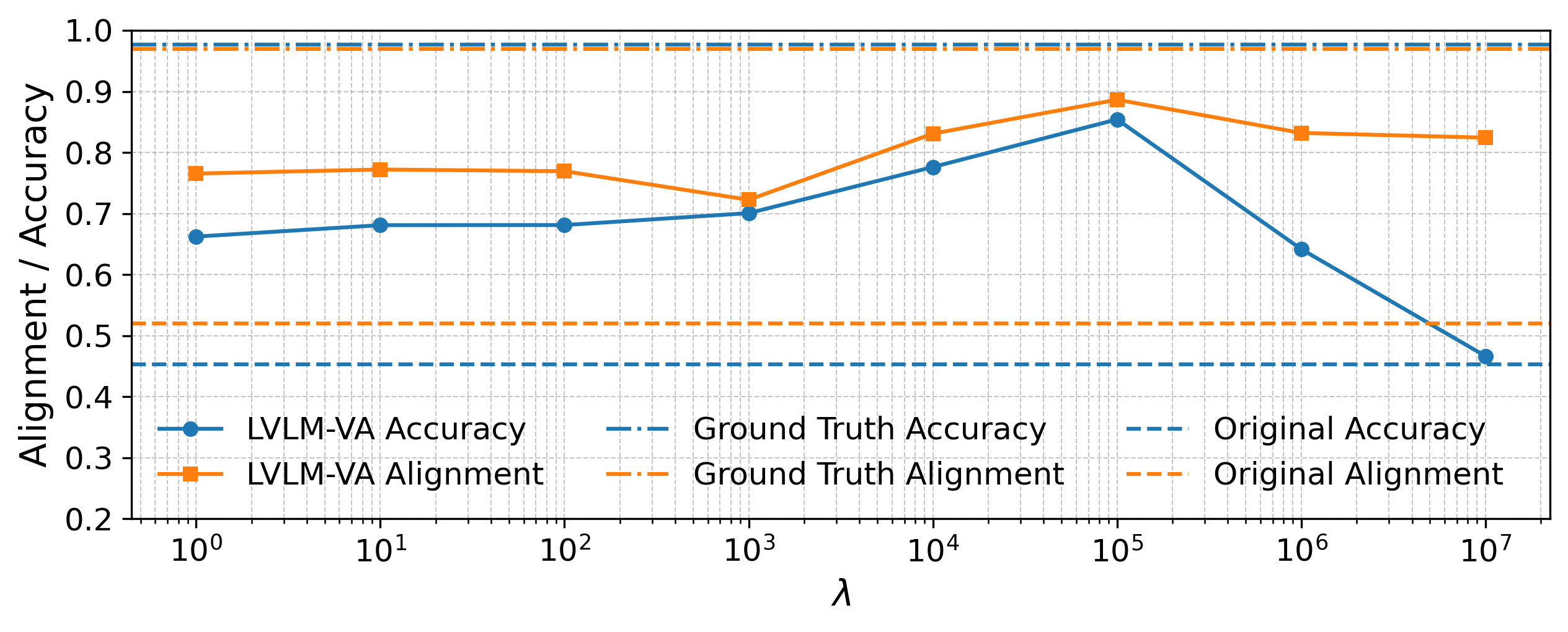}
    \caption{Alignment and accuracy on the test set across different $\lambda$ values weighing the influence of the RRR loss term during alignment. Alignment as well as accuracy rise with increase in $\lambda$ until $10^5$ where the accuracy drops substantially. At $\lambda = 10^5$ both metrics are close to the upper bound reached when using the ground truth correction masks.}
    \label{fig:lambda_mnist}
\end{figure}
To also quantitatively evaluate the benefit of our method in this setting, we use the intuitive lower and upper bounds for human involvement as a baseline. The least human involvement is given by not aligning the model at all and only using the simple target labels during the original training process. In contrast, using instance-wise human-generated expert masks $A^{(GT)}$ in combination with the RRR loss requires significant manual effort if not automatically generated as in this synthetic setting. The availability of information about the location of the spurious features in this scenario allows us to also quantitatively evaluate the alignment of the model. For this, we introduce an alignment metric adapted from \cite{kohlbrenner2020towards, koebler_teoe} between the ground truth masks $A^{(GT)}$, i.e., the artificial decoys in the corner, and the absolute explanation maps after alignment $|\Phi (x,y,f)|$. This metric measures the fraction of attribution mass that lies outside the ground-truth spurious region and therefore on the relevant digit features. For $N_t$ test samples it is defined as
\begin{equation*}
    \mu_{Align}
    =
    1 - \frac{1}{N_t}\sum_{n=1}^{N_t}
    \frac{\sum_{i=1}^{M}A^{(GT)}_{n,i}|\Phi_i(x_n,y_n,f)|}
    {\sum_{i=1}^{M}|\Phi_i(x_n,y_n,f)|} \; .
\end{equation*}
As shown in \cref{fig:lambda_mnist}, our LVLM-VA approach improves both performance and model alignment. This effect is emphasized by increasing the influence of the RRR loss up to $\lambda=10^5$. However, at this point, performance drops drastically as the cross-entropy loss becomes negligible. LVLM-VA achieves values approaching those obtained using ground truth instance-wise feedback within the RRR loss, which requires substantial manual labelling effort.

\subsection{Mitigating Spurious Correlations in Real-World Medical Datasets}
\begin{figure}[h]
    \centering
    \includegraphics[width=0.9\columnwidth]{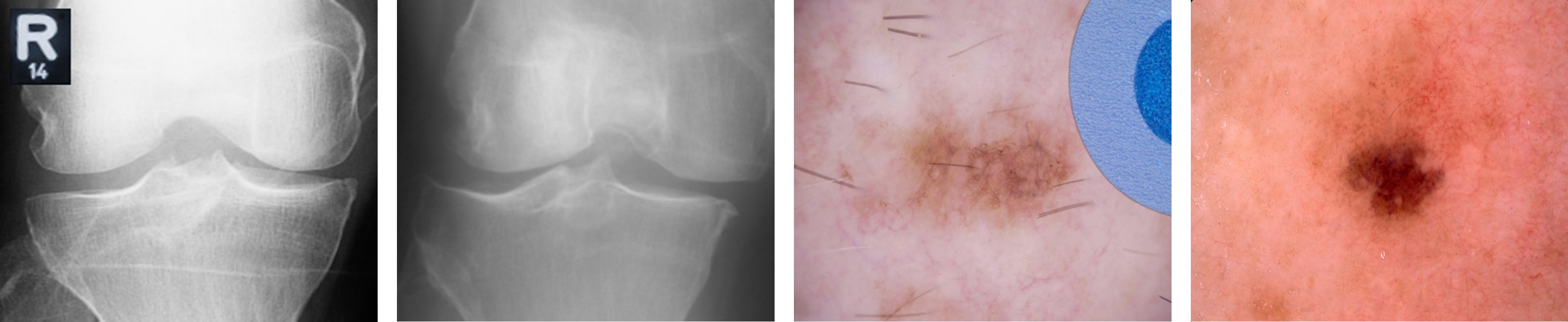}
    \caption{Prototypical images for the medical datasets. Some of the knee radiograph images (left) include spurious hospital tags whereas the skin lesion images (right) include colored bandages. Both of those spurious features might be learned as simple shortcuts compared to the complex original classification task.}
    \label{fig:med_datasets}
\end{figure}

\begin{figure}[h]
    \centering
    \includegraphics[width=1.0\columnwidth]{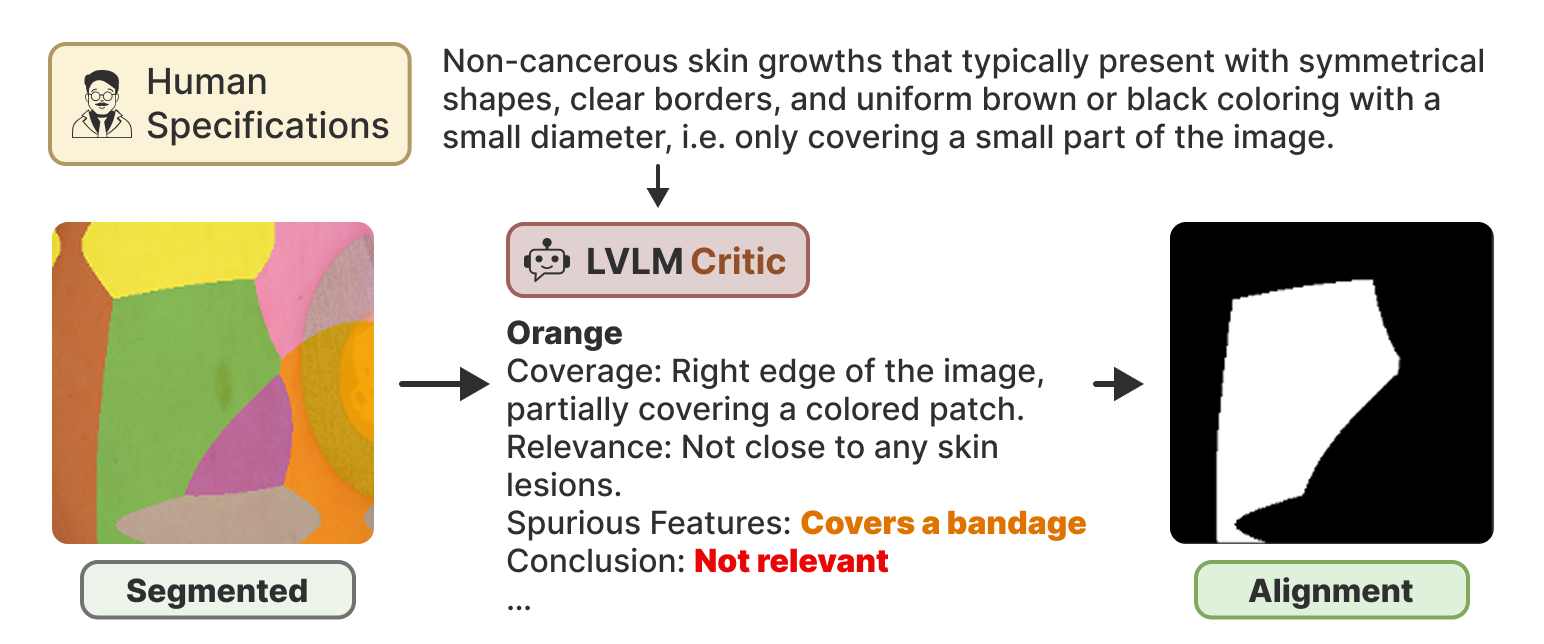}
    \caption{Prototypical shortcut detection results for the \textit{skin lesion dataset}. Only the green cluster in the segmented input image covers the actual skin lesion. This enables the LVLM-Critic, which is informed by the human class description, to deem all other clusters as irrelevant. This produces a correction mask $A$, where gradients will be penalized except in the green segment during the fine-tuning step.}
    \label{fig:skin_plot}
\end{figure}

\begin{figure}[h]
    \centering
    \includegraphics[width=\columnwidth]{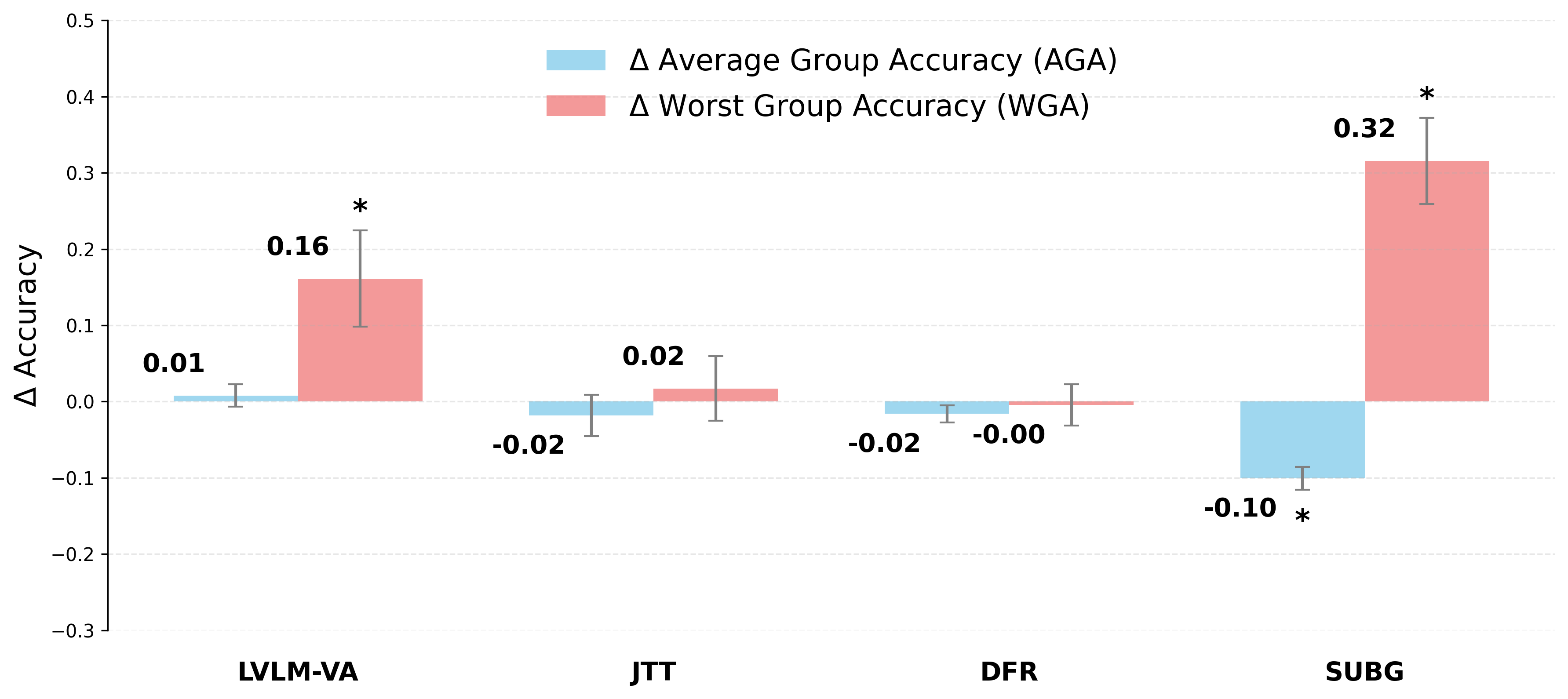}
    \caption{Change in Average Group Accuracy (AGA) and Worst Group Accuracy (WGA) relative to the original model after shortcut mitigation on the \textit{knee radiographs dataset}. Results are averaged over seven random seeds (mean ± std). LVLM-VA is the only method which increases the WGA whilst maintaining overall accuracy. (*: Wilcoxon Signed-Rank Test $p<0.05$)}
    \label{fig:acc_knee}
\end{figure}

\begin{figure}[h]
    \centering
    \includegraphics[width=\columnwidth]{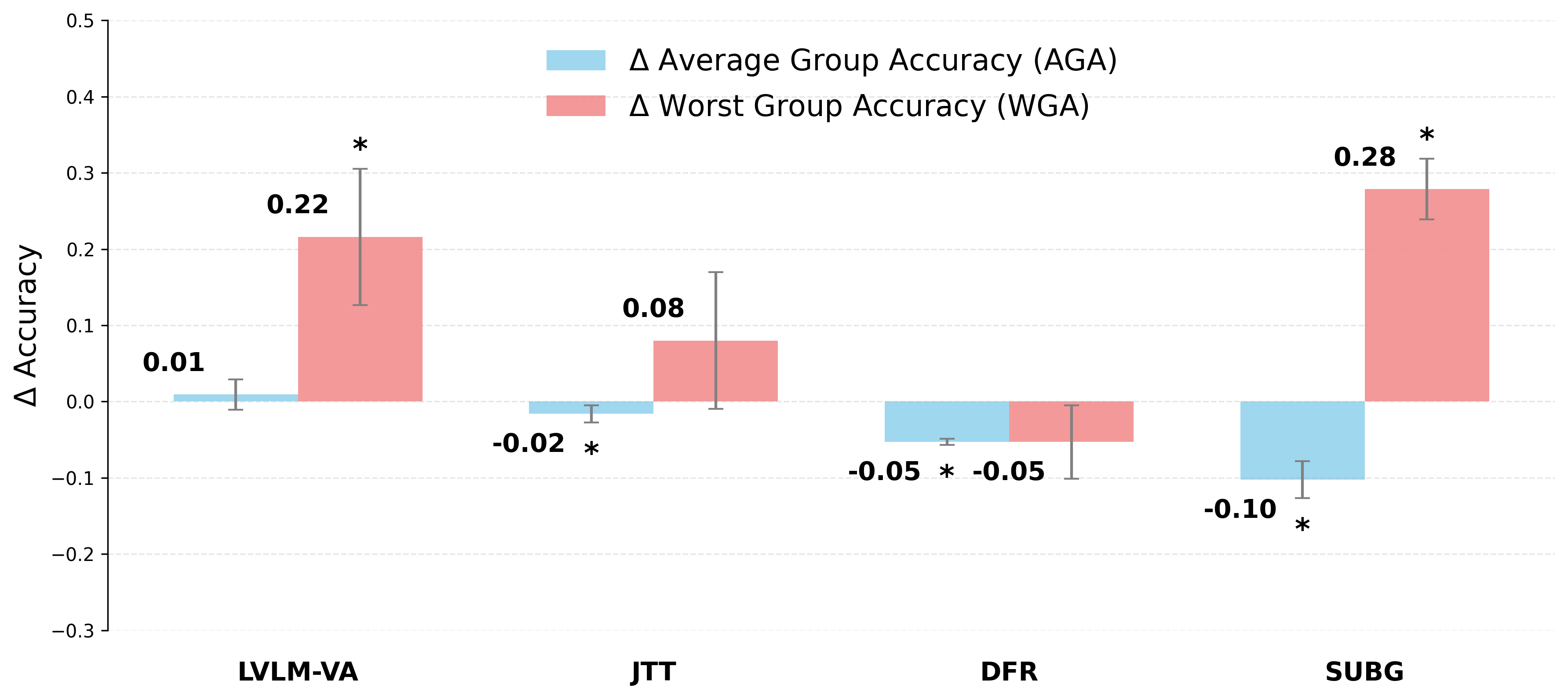}
    \caption{Change in Average Group Accuracy (AGA) and Worst Group Accuracy (WGA) relative to the original model after shortcut mitigation on the \textit{skin lesion dataset}. Results are averaged over seven random seeds (mean ± std). LVLM-VA is the only method which increases the WGA whilst maintaining overall accuracy. (*: Wilcoxon Signed-Rank Test $p<0.05$).}
    \label{fig:acc_skin}
\end{figure}

In the following two experiments, we investigate the effectiveness of LVLM-VA to mitigate learned shortcuts in two real-world medical datasets shown in \cref{fig:med_datasets}.
In the medical domain, it is of vital importance to ensure consistent model performance despite often limited data quality and quantity. Therefore, the model's performance should be independent of any spurious elements introduced in a subset of the available images that could lead to biases for any given (potentially protected) group. For this reason, we measure the effectiveness of our approach by its ability to improve the accuracy of the worst performing group (a standard metric for shortcut mitigation strategies \cite{kirichenkolast, zheng2024learning}) whilst at least maintaining overall accuracy. As the manual curation of medical datasets is very laborious and the time of medical experts is limited, LVLM-VA can help to automatically align the model with high-level expert input.\\
\textbf{Experimental Setting:}
First, we evaluate our method on the \textit{International Skin Imaging Collaboration (ISIC) skin lesion dataset} \cite{8363547}, which is also used in multiple other shortcut mitigation works \cite{le2023last, diagnostics12010040, bekkemoen2021correcting}. This dataset contains images of real skin lesions, which are either benign or malignant tumors. Some of the images contain bandages of different colours, which are located randomly next to the skin lesions. Our training set consists of 1,800 samples per class. For the benign class, there are an equal number of images containing and not containing coloured bandages. In contrast, only ten images in the malignant class contain one or multiple bandages. This difference in the occurrence of bandages across the classes renders them a spurious feature that can easily be learned as a shortcut during the initial training phase. In the test set, bandages occur equally frequently, leading to low accuracy for images of malignant lesions containing bandages. For our LVLM-VA approach, we use an alignment dataset of size $N = 1024$ and provide short human descriptions for both classes. \newline
Secondly, we use a \textit{knee osteoarthritis radiograph dataset} \cite{CHEN201984, kuhn2025efficient} that includes images depicting various stages of osteoarthritis.
We evaluate the binary classification task of distinguishing between 'no' and 'moderate' osteoarthritis. Similar as done by \citet{DeGravexray}, shortcuts are added in the form of hospital tags specified as 'L' or 'R' to the edges of the images, indicating the right or left knee. The occurrence of these tags in the training set is class-dependent, with 50\% of healthy knees and only 2.5\% of arthritic knees co-occurring with hospital tags. In total, there are 1,000 training samples. Furthermore, there are 400 test samples, distributed equally across the groups. As expected, the hospital tags lead to spurious shortcuts, resulting in low group accuracy for arthritic knee images with hospital tags. The alignment dataset includes $N = 256$ samples, and we provide short human descriptions for both classes.\\
For both experiments, we use a ResNet50 model \cite{he2016deep} and set $\lambda = 1$. We set the number of clusters $J$ to seven. The performance is stable for reasonably large $J$ (see Appendix).\\
We benchmark LVLM-VA against the three commonly applied shortcut mitigation methods: sub-sampling groups (SUBG) \cite{idrissi2022simple}, Deep Feature Reweighting \cite{kirichenkolast} (DFR), and Just Train Twice \cite{liu2021just} (JTT).\\
\textbf{Results:}
\cref{fig:skin_plot} illustrates the initial stage of identifying potential spurious features and generating the corresponding correction maps $A$ for the skin dataset. The proposed clustering approach, based on model explanations combined with PPEPS-WGM, generates one or more clusters covering irrelevant features without substantially overlapping the core features relevant for classification (i.e. the knee or skin lesion). The provided human descriptions help the LVLM-Critic identify which parts of the image should be considered core features, given a particular class. Based on this information, the LVLM-Critic provides intermediate assessments, enforced by the chain-of-thought prompt, as well as a clear, combined statement about the relevance of the considered cluster. This enables the Judge to efficiently generate a structured binary verdict for each cluster. These verdicts are then translated into the displayed correction masks. For knee radiographs, where the LVLM is unable to judge and is not intended to identify which part of the knee is relevant, it only declares the cluster that actually includes the spurious hospital tag as irrelevant (see \cref{fig:exp}). In contrast, for the ISIC dataset, where it is clear that clusters not covering the skin lesion should be irrelevant for the classification, the LVLM also labels clusters as irrelevant even though they do not include a bandage (see \cref{fig:skin_plot}). This approach enables spurious features that are neither described by humans nor detected by the critic to be removed (e.g. slight shadows or reflections) without eliminating potentially important core features and sacrificing overall performance.
In \cref{fig:acc_knee} and \cref{fig:acc_skin}, the change in the model's average group / overall accuracy ($\Delta$AGA) and the worst group accuracy ($\Delta$WGA) before and after the shortcut mitigation step for LVLM-VA is compared to that of the three baselines. Across both experiments, LVLM-VA enables a significant improvement in the worst group accuracy without diminishing the overall accuracy. The two step JTT approach enables a slight improvement in WGA for the skin dataset, but does not consistently improve performance for both datasets. Subsampling groups (SUBG) improves WGA, even outperforming LVLM-VA on the knee dataset, but significantly sacrifices overall accuracy, rendering it invalid for most applications. Deep Feature Reweighting (DFR) was not beneficial in either experiment.
The experiments on medical datasets demonstrate that LVLM-VA is the most effective way of mitigating group biases caused by spurious features, without requiring to specify which samples are affected by shortcuts. These instance-wise group labels are significantly more labor intensive to generate than human descriptions at the class level.

\section{Ablations and Discussion}
In this section, we present a user study to evaluate the effect of LVLM-VA on the perceived alignment of vision models. In addition, we conducted several experiments that examine the choice of sub-components of LVLM-VA and further discuss the limitations and scope.\\
\textbf{User Study on Alignment:}
The conducted user study includes 18 participants with a data science background and familiarity with medical imaging. They assess three aspects across 26 questions covering both medical datasets:
(1) \textit{Cluster relevance}: Participants are asked to select irrelevant/spurious clusters given the original and the segmented image. They agreed with the LVLM-selected clusters in 88\% of cases, which is in line with the measured verdict accuracy of 87\% on the entire alignment set.
(2) \textit{Critic reasoning}: Participants are presented with the Critic’s natural-language argumentation in addition to the previous images and tasked to assess its correctness. They agreed with the LVLM's argumentation in 87\% of cases.
(3) \textit{Explanation alignment}: The participants are shown the target model explanations before and after the alignment step. Participants conclude that the post-alignment explanations are better aligned with their expectations in 86\% of cases. This is consistent with the alignment analysis on DecoyMNIST (Fig. \ref{fig:exp_MNIST} \& Fig. \ref{fig:lambda_mnist}). Together, these results underline that LVLM-VA can significantly contribute to a better alignment of vision models with human expectations.\\
\textbf{LVLM Choice:} Our method leverages recent advances in rapidly evolving LVLMs. During the main development phase of this work, GPT-4o was used. However, Table \ref{tab:LVLMs} shows that newer models outperform older ones in verdict accuracy, which measures whether a cluster flagged as spurious actually contains part of the spurious feature. This higher verdict accuracy leads to an increased WGA, while at the same time, newer models continue to decrease in cost, thereby steadily improving the accessibility of our method.\\
\begin{table}[ht]
\centering
\caption{Benchmark verdict accuracy (knee) and $\Delta$WGA for different Critic/Judge LVLMs. Costs in USD per one million input tokens (Nov. 2025). (*: Wilcoxon Signed-Rank Test for the improvement in WGA w.r.t. the original model across 7 seeds $p<0.05$)}
\label{tab:LVLMs}
\begin{tabular}{lccc}
\toprule
\textbf{Model} & \textbf{Cost} & \textbf{Verdict Acc.} & \textbf{$\Delta$WGA} \\
\midrule
GPT-4o (used) & 2.50 & 0.87 & $0.16 \pm 0.06$* \\
GPT-5         & 1.25 & 1.00 & $0.20 \pm 0.09$* \\
GPT-4o-mini   & 0.15 & 0.42 & $0.09 \pm 0.02$* \\
\bottomrule
\end{tabular}
\end{table}
\textbf{Sampling Strategy:} We further reduced the LVLM costs by introducing a low-entropy sampling strategy, which preferentially applies the Critic \& Judge only to samples containing spurious features. On the knee dataset, this strategy produces an alignment set in which 56\% of images contain spurious features, compared to only 25\% under random sampling and 2\% when sampling based on high entropy.\\
\textbf{Segmentation Method:} To make the most effective use of the limited alignment set, we introduced PPEPS-WGM as a segmentation method that targets clusters with high positive attribution density, rather than segmenting the underlying image content directly. To assess the benefit of PPEPS-WGM, we compare it to segmenting the input images with a Segment Anything Model (SAM) \cite{kirillov2023segany} as done by \citet{yang2023set}. Table \ref{tab:gmm_sam_wagdiff} shows that, although the verdict accuracy of the two approaches is similar, the overall improvement in $\Delta$WGA achieved with SAM is smaller. This is because SAM frequently groups the spurious feature together with the relevant knee structures into a single segment. In contrast, PPEPS-WGM more effectively isolates the spurious feature by clustering spatially separated positive attribution.
\begin{table}[ht]
    \centering
    \caption{Verdict accuracy and corresponding increase in $\Delta$WGA (knee) for SAM\_VIT\_B model (SAM) and PPEPS-WGM. (*: Wilcoxon Signed-Rank Test for the improvement in WGA w.r.t. the original model across 7 seeds $p<0.05$)}
    \begin{tabular}{lcc}
        \toprule
        \textbf{Method} & \textbf{Verdict Acc.} & \textbf{$\Delta$ WGA} \\
        \midrule
        PPEPS-WGM & $0.87$ & $0.16 \pm 0.06$* \\
        SAM & $0.87$ & $0.11 \pm 0.04$* \\
        \bottomrule
    \end{tabular}
    \label{tab:gmm_sam_wagdiff}
\end{table}

\textbf{Limitations:} Although LVLM-VA removes the need for fine-grained annotations, it relies on class-level descriptions provided by domain experts. In some cases, these descriptions may be difficult to formalize, as experts have rather learned those patterns intuitively. Furthermore, the distinction between core and spurious features is not always clear or spatially separable.
We address this issue by having differing degrees of intervention between the skin lesion and knee radiograph datasets. While the LVLM intervenes on clear spurious features in the case of the synthetic decoy and knee dataset, it rather focuses on preserving clearly described core features in the form of skin lesions for the ISIC dataset. Being able to explicitly describe either core or spurious features applies to most real-world use-cases making our method applicable to a wide variety of settings.

\section{Conclusion}
We have proposed a novel approach, LVLM-Aided Visual Alignment (LVLM-VA), to correct spurious correlations and increase the overall and worst group accuracy of small, task-specific vision models. LVLM-VA translates model behavior into natural language and incorporates human descriptions at the class level via instance-wise critique into the model. This provides an efficient human-centered interface for aligning vision models with domain knowledge, eliminating the need for expensive fine-grained feedback or group labels. LVLM-VA fosters synergies between generative AI models and more explainable, established discriminative approaches.

{
    \small
    \bibliographystyle{ieeenat_fullname}
    \bibliography{main}

@String(CVPR= {IEEE Conf. Comput. Vis. Pattern Recog.})

@String(NIPS= {Adv. Neural Inform. Process. Syst.})

@String(ICLR = {Int. Conf. Learn. Represent.})

@String(IJCAI = {IJCAI})

@String(AAAI = {AAAI})

@String(CVPR  = {CVPR})

@String(NIPS  = {NeurIPS})

@String(ICLR  = {ICLR})

@article{DeGravexray,
	title = {{AI} for radiographic {COVID}-19 detection selects shortcuts over signal},
	volume = {3},
	issn = {2522-5839},
	url = {https://doi.org/10.1038/s42256-021-00338-7},
	doi = {10.1038/s42256-021-00338-7},
	number = {7},
	journal = {Nature Machine Intelligence},
	author = {DeGrave, Alex J. and Janizek, Joseph D. and Lee, Su-In},
	month = jul,
	year = {2021},
	pages = {610--619},
}

@article{stammerlearning,
  title={Learning by self-explaining},
  author={Stammer, Wolfgang and Friedrich, Felix and Steinmann, David and Brack, Manuel and Shindo, Hikaru and Kersting, Kristian},
  journal={arXiv preprint arXiv:2309.08395},
  year={2023}
}

@article{von2021informed,
  title={Informed machine learning--a taxonomy and survey of integrating prior knowledge into learning systems},
  author={Von Rueden, Laura and Mayer, Sebastian and Beckh, Katharina and Georgiev, Bogdan and Giesselbach, Sven and Heese, Raoul and Kirsch, Birgit and Pfrommer, Julius and Pick, Annika and Ramamurthy, Rajkumar and others},
  journal={IEEE Transactions on Knowledge and Data Engineering},
  volume={35},
  number={1},
  pages={614--633},
  year={2021},
  publisher={IEEE}
}

@article{rueckel_conf_med,
author = {Rueckel, Johannes and Trappmann, Lena and Schachtner, Balthasar and Wesp, Philipp and Hoppe, Boj and Fink, Nicola and Ricke, Jens and Dinkel, julien and Ingrisch, Michael and Sabel, Bastian},
year = {2020},
month = {07},
pages = {},
title = {Impact of Confounding Thoracic Tubes and Pleural Dehiscence Extent on Artificial Intelligence Pneumothorax Detection in Chest Radiographs},
volume = {Publish Ahead of Print},
journal = {Investigative Radiology},
doi = {10.1097/RLI.0000000000000707}
}

@article{lapuschkin_clever_hans,
author = {Lapuschkin, Sebastian and Wäldchen, Stephan and Binder, Alexander and Montavon, Grégoire and Samek, Wojciech and Müller, Klaus-Robert},
year = {2019},
month = {03},
pages = {},
title = {Unmasking Clever Hans Predictors and Assessing What Machines Really Learn},
volume = {10},
journal = {Nature Communications},
doi = {10.1038/s41467-019-08987-4}
}

@InProceedings{Guan_2024_CVPR,
    author    = {Guan, Tianrui and Liu, Fuxiao and Wu, Xiyang and Xian, Ruiqi and Li, Zongxia and Liu, Xiaoyu and Wang, Xijun and Chen, Lichang and Huang, Furong and Yacoob, Yaser and Manocha, Dinesh and Zhou, Tianyi},
    title     = {HallusionBench: An Advanced Diagnostic Suite for Entangled Language Hallucination and Visual Illusion in Large Vision-Language Models},
    booktitle = {Proceedings of the IEEE/CVF Conference on Computer Vision and Pattern Recognition (CVPR)},
    month     = {June},
    year      = {2024},
    pages     = {14375-14385}
}

@inproceedings{yang2021causal,
  title={Causal attention for vision-language tasks},
  author={Yang, Xu and Zhang, Hanwang and Qi, Guojun and Cai, Jianfei},
  booktitle={Proceedings of the IEEE/CVF conference on computer vision and pattern recognition},
  pages={9847--9857},
  year={2021}
}

@inproceedings{decker2023thousand,
  title={The thousand faces of explainable AI along the machine learning life cycle: industrial reality and current state of research},
  author={Decker, Thomas and Gross, Ralf and Koebler, Alexander and Lebacher, Michael and Schnitzer, Ronald and Weber, Stefan H},
  booktitle={International Conference on Human-Computer Interaction},
  pages={184--208},
  year={2023},
  organization={Springer}
}

@inproceedings{gu2024anomalygpt,
  title={Anomalygpt: Detecting industrial anomalies using large vision-language models},
  author={Gu, Zhaopeng and Zhu, Bingke and Zhu, Guibo and Chen, Yingying and Tang, Ming and Wang, Jinqiao},
  booktitle={Proceedings of the AAAI Conference on Artificial Intelligence},
  volume={38},
  number={3},
  pages={1932--1940},
  year={2024}
}

@article{shen2024towards,
  title={Towards Bidirectional Human-AI Alignment: A Systematic Review for Clarifications, Framework, and Future Directions},
  author={Shen, Hua and Knearem, Tiffany and Ghosh, Reshmi and Alkiek, Kenan and Krishna, Kundan and Liu, Yachuan and Ma, Ziqiao and Petridis, Savvas and Peng, Yi-Hao and Qiwei, Li and others},
  journal={arXiv preprint arXiv:2406.09264},
  year={2024}
}

@inproceedings{DeepShap,
author = {Lundberg, Scott M. and Lee, Su-In},
title = {A unified approach to interpreting model predictions},
year = {2017},
isbn = {9781510860964},
publisher = {Curran Associates Inc.},
address = {Red Hook, NY, USA},
booktitle = {Proceedings of the 31st International Conference on Neural Information Processing Systems},
pages = {4768–4777},
numpages = {10},
location = {Long Beach, California, USA},
series = {NIPS'17}
}

@article { lecun-98,
original =    "orig/lecun-98.ps.gz",
author = 	"LeCun, Y. and Bottou, L. and Bengio, Y. and Haffner, P.",
title = 	"Gradient-Based Learning Applied to Document Recognition",
journal =	"Proceedings of the IEEE",
month =         "November",
volume =        "86",
number =        "11",
pages =         "2278-2324",
year =		1998
}

@article{brown2020language,
  title={Language models are few-shot learners},
  author={Brown, Tom and Mann, Benjamin and Ryder, Nick and Subbiah, Melanie and Kaplan, Jared D and Dhariwal, Prafulla and Neelakantan, Arvind and Shyam, Pranav and Sastry, Girish and Askell, Amanda and others},
  journal={Advances in neural information processing systems},
  volume={33},
  pages={1877--1901},
  year={2020}
}

@article{yang2023set,
  title={Set-of-mark prompting unleashes extraordinary visual grounding in gpt-4v},
  author={Yang, Jianwei and Zhang, Hao and Li, Feng and Zou, Xueyan and Li, Chunyuan and Gao, Jianfeng},
  journal={arXiv preprint arXiv:2310.11441},
  year={2023}
}

@inproceedings{ross2017right,
author = {Ross, Andrew Slavin and Hughes, Michael C. and Doshi-Velez, Finale},
title = {Right for the right reasons: training differentiable models by constraining their explanations},
year = {2017},
isbn = {9780999241103},
publisher = {AAAI Press},
booktitle = {Proceedings of the 26th International Joint Conference on Artificial Intelligence},
pages = {2662–2670},
numpages = {9},
location = {Melbourne, Australia},
series = {IJCAI'17}
}

@inproceedings{slany2022caipi,
  title={CAIPI in practice: towards explainable interactive medical image classification},
  author={Slany, Emanuel and Ott, Yannik and Scheele, Stephan and Paulus, Jan and Schmid, Ute},
  booktitle={IFIP International Conference on Artificial Intelligence Applications and Innovations},
  pages={389--400},
  year={2022},
  organization={Springer}
}

@article{schramowski2020making,
  title={Making deep neural networks right for the right scientific reasons by interacting with their explanations},
  author={Schramowski, Patrick and Stammer, Wolfgang and Teso, Stefano and Brugger, Anna and Herbert, Franziska and Shao, Xiaoting and Luigs, Hans-Georg and Mahlein, Anne-Katrin and Kersting, Kristian},
  journal={Nature Machine Intelligence},
  volume={2},
  number={8},
  pages={476--486},
  year={2020},
  publisher={Nature Publishing Group UK London}
}

@inproceedings{teso_xil,
author = {Teso, Stefano and Kersting, Kristian},
title = {Explanatory Interactive Machine Learning},
year = {2019},
isbn = {9781450363242},
publisher = {Association for Computing Machinery},
address = {New York, NY, USA},
url = {https://doi.org/10.1145/3306618.3314293},
doi = {10.1145/3306618.3314293},
booktitle = {Proceedings of the 2019 AAAI/ACM Conference on AI, Ethics, and Society},
pages = {239–245},
numpages = {7},
location = {Honolulu, HI, USA},
series = {AIES '19}
}

@inproceedings{kohlbrenner2020towards,
  title={Towards best practice in explaining neural network decisions with LRP},
  author={Kohlbrenner, Maximilian and Bauer, Alexander and Nakajima, Shinichi and Binder, Alexander and Samek, Wojciech and Lapuschkin, Sebastian},
  booktitle={2020 International Joint Conference on Neural Networks (IJCNN)},
  pages={1--7},
  year={2020},
  organization={IEEE}
}

@inproceedings{koebler_teoe,
author = {Koebler, Alexander and Greisinger, Christian and Paulus, Jan and Thon, Ingo and Buettner, Florian},
title = {Through the Eyes of the Expert: Aligning Human and Machine Attention for Industrial AI},
year = {2024},
isbn = {978-3-031-60613-7},
publisher = {Springer-Verlag},
address = {Berlin, Heidelberg},
url = {https://doi.org/10.1007/978-3-031-60611-3_28},
doi = {10.1007/978-3-031-60611-3_28},
booktitle = {Artificial Intelligence in HCI: 5th International Conference, AI-HCI 2024, Held as Part of the 26th HCI International Conference, HCII 2024, Washington, DC, USA, June 29–July 4, 2024, Proceedings, Part II},
pages = {407–423},
numpages = {17},
location = {Washington DC, USA}
}

@inproceedings{hermannfoundations,
  author       = {Katherine L. Hermann and
                  Hossein Mobahi and
                  Thomas Fel and
                  Michael Curtis Mozer},
  title        = {On the Foundations of Shortcut Learning},
  booktitle    = {The Twelfth International Conference on Learning Representations,
                  {ICLR} 2024, Vienna, Austria, May 7-11, 2024},
  publisher    = {OpenReview.net},
  year         = {2024},
  url          = {https://openreview.net/forum?id=Tj3xLVuE9f},
  bibsource    = {dblp computer science bibliography, https://dblp.org}
}

@article{wei2022chain,
  title={Chain-of-thought prompting elicits reasoning in large language models},
  author={Wei, Jason and Wang, Xuezhi and Schuurmans, Dale and Bosma, Maarten and Xia, Fei and Chi, Ed and Le, Quoc V and Zhou, Denny and others},
  journal={Advances in neural information processing systems},
  volume={35},
  pages={24824--24837},
  year={2022}
}

@inproceedings{idrissi2022simple,
  title={Simple data balancing achieves competitive worst-group-accuracy},
  author={Idrissi, Badr Youbi and Arjovsky, Martin and Pezeshki, Mohammad and Lopez-Paz, David},
  booktitle={Conference on Causal Learning and Reasoning},
  pages={336--351},
  year={2022},
  organization={PMLR}
}

@inproceedings{liu2021just,
  title={Just train twice: Improving group robustness without training group information},
  author={Liu, Evan Z and Haghgoo, Behzad and Chen, Annie S and Raghunathan, Aditi and Koh, Pang Wei and Sagawa, Shiori and Liang, Percy and Finn, Chelsea},
  booktitle={International Conference on Machine Learning},
  pages={6781--6792},
  year={2021},
  organization={PMLR}
}

@article{kirichenkolast,
  title={Last Layer Re-Training is Sufficient for Robustness to Spurious Correlations},
  author={Kirichenko, Polina and Izmailov, Pavel and Gordon Wilson, Andrew},
  journal={ICLR 2023},
  year={2023}
}

@article{le2023last,
  title={Is Last Layer Re-Training Truly Sufficient for Robustness to Spurious Correlations?},
  author={Le, Phuong Quynh and Schl{\"o}tterer, J{\"o}rg and Seifert, Christin},
  journal={arXiv preprint arXiv:2308.00473},
  year={2023}
}

@article{bekkemoen2021correcting,
  title={Correcting classification: A bayesian framework using explanation feedback to improve classification abilities},
  author={Bekkemoen, Yanzhe and Langseth, Helge},
  journal={arXiv preprint arXiv:2105.02653},
  year={2021}
}

@inproceedings{he2016deep,
  title={Deep residual learning for image recognition},
  author={He, Kaiming and Zhang, Xiangyu and Ren, Shaoqing and Sun, Jian},
  booktitle={Proceedings of the IEEE conference on computer vision and pattern recognition},
  pages={770--778},
  year={2016}
}

@Article{diagnostics12010040,
AUTHOR = {Nauta, Meike and Walsh, Ricky and Dubowski, Adam and Seifert, Christin},
TITLE = {Uncovering and Correcting Shortcut Learning in Machine Learning Models for Skin Cancer Diagnosis},
JOURNAL = {Diagnostics},
VOLUME = {12},
YEAR = {2022},
NUMBER = {1},
ARTICLE-NUMBER = {40},
URL = {https://www.mdpi.com/2075-4418/12/1/40},
PubMedID = {35054207},
ISSN = {2075-4418},
DOI = {10.3390/diagnostics12010040}
}

@INPROCEEDINGS{8363547,
  author={Codella, Noel C. F. and Gutman, David and Celebi, M. Emre and Helba, Brian and Marchetti, Michael A. and Dusza, Stephen W. and Kalloo, Aadi and Liopyris, Konstantinos and Mishra, Nabin and Kittler, Harald and Halpern, Allan},
  booktitle={2018 IEEE 15th International Symposium on Biomedical Imaging (ISBI 2018)}, 
  title={Skin lesion analysis toward melanoma detection: A challenge at the 2017 International symposium on biomedical imaging (ISBI), hosted by the international skin imaging collaboration (ISIC)}, 
  year={2018},
  volume={},
  number={},
  pages={168-172},
  doi={10.1109/ISBI.2018.8363547}}

@article{CHEN201984,
title = {Fully automatic knee osteoarthritis severity grading using deep neural networks with a novel ordinal loss},
journal = {Computerized Medical Imaging and Graphics},
volume = {75},
pages = {84-92},
year = {2019},
issn = {0895-6111},
doi = {https://doi.org/10.1016/j.compmedimag.2019.06.002},
url = {https://www.sciencedirect.com/science/article/pii/S0895611118304956},
author = {Pingjun Chen and Linlin Gao and Xiaoshuang Shi and Kyle Allen and Lin Yang}
}

@misc{kokhlikyan2020captum,
    title={Captum: A unified and generic model interpretability library for PyTorch},
    author={Narine Kokhlikyan and Vivek Miglani and Miguel Martin and Edward Wang and Bilal Alsallakh and Jonathan Reynolds and Alexander Melnikov and Natalia Kliushkina and Carlos Araya and Siqi Yan and Orion Reblitz-Richardson},
    year={2020},
    eprint={2009.07896},
    archivePrefix={arXiv},
    primaryClass={cs.LG}
}

@article{kingma2014adam,
  title={Adam: A method for stochastic optimization},
  author={Kingma, Diederik P and Ba, Jimmy},
  journal={arXiv preprint arXiv:1412.6980},
  year={2014}
}

@inproceedings{zheng2024learning,
  title={Learning robust classifiers with self-guided spurious correlation mitigation},
  author={Zheng, Guangtao and Ye, Wenqian and Zhang, Aidong},
  booktitle={Proceedings of the Thirty-Third International Joint Conference on Artificial Intelligence},
  pages={5599--5607},
  year={2024}
}

@inproceedings{kirillov2023segany,
  title={Segment anything},
  author={Kirillov, Alexander and Mintun, Eric and Ravi, Nikhila and Mao, Hanzi and Rolland, Chloe and Gustafson, Laura and Xiao, Tete and Whitehead, Spencer and Berg, Alexander C and Lo, Wan-Yen and others},
  booktitle={Proceedings of the IEEE/CVF international conference on computer vision},
  pages={4015--4026},
  year={2023}
}

@inproceedings{kuhn2025efficient,
  title={Efficient unsupervised shortcut learning detection and mitigation in transformers},
  author={Kuhn, Lukas and Sadiya, Sari and Schl{\"o}tterer, J{\"o}rg and Buettner, Florian and Seifert, Christin and Roig, Gemma},
  booktitle={Proceedings of the IEEE/CVF International Conference on Computer Vision},
  pages={2217--2226},
  year={2025}
}

@misc{meta2024llama4,
  title        = {Introducing {LLaMA} 4: Advancing Multimodal Intelligence},
  author       = {{Meta AI}},
  year         = {2024},
  howpublished = {\url{https://ai.meta.com/blog/llama-4-multimodal-intelligence/}}
}

@article{yang2025qwen3,
  title={Qwen3 technical report},
  author={Yang, An and Li, Anfeng and Yang, Baosong and Zhang, Beichen and Hui, Binyuan and Zheng, Bo and Yu, Bowen and Gao, Chang and Huang, Chengen and Lv, Chenxu and others},
  journal={arXiv preprint arXiv:2505.09388},
  year={2025}
}

@article{di2024medmnist,
  title={Medmnist-c: Comprehensive benchmark and improved classifier robustness by simulating realistic image corruptions},
  author={Di Salvo, Francesco and Doerrich, Sebastian and Ledig, Christian},
  journal={arXiv preprint arXiv:2406.17536},
  year={2024}
}
}
\setcounter{page}{1}
\appendix
\maketitlesupplementary

\section{Additional Experimental Results}
In this section, we present ablations for the use of open-source LVLMs, the selection of the number of clusters $J$ in PPEPS-WGM, and the effect of LVLM-aided alignment on the robustness of vision models to global perturbations. Further, we provide additional insights on the group accuracy gains on the medical datasets and on the effect of the alignment on the embedding space for DecoyMNIST.

\subsection{Open-Source LVLMs} In Table \ref{tab:LVLMs}, we compare the verdict accuracy of the OpenAI GPT models used in the main paper with recent open-source alternatives. Verdict accuracy is the proportion of clusters that overlap spurious features and are then actually correctly classified as spurious by the Critic \& Judge pair. As open-source models, we evaluate Llama-4-Scout (\textit{Llama-4-Scout-17B-16E-Instruct}) \cite{meta2024llama4} as both Critic and Judge, and Qwen3 models \cite{yang2025qwen3} using \textit{Qwen3-VL-235B-A22B-Thinking} as the Critic and \textit{Qwen3-235B-A22B-Instruct-2507} as the Judge. While the Qwen3 models under-perform the larger OpenAI models, Llama-4 achieves a verdict accuracy comparable to GPT-4o. This demonstrates that our method does not rely on any specific proprietary LVLM and can also be effectively applied with open-source models. Together with the other measures discussed in the main paper, this further enhances the accessibility of LVLM-VA.
\begin{table}[ht]
\centering
\caption{Comparison of the verdict accuracy for different Critic \& Judge pairs including open source models on the knee alignment samples.}
\label{tab:LVLMs}
\begin{tabular}{lccccc}
\toprule
\textbf{GPT-5} & \textbf{Llama4} & \textbf{GPT-4o} & \textbf{Qwen3} & \textbf{GPT-4o-mini} \\
\midrule
1.00 & 0.88 & 0.87 & 0.46 & 0.42 \\
\bottomrule
\end{tabular}
\end{table}
\subsection{Number of Clusters} 
\begin{figure}[h]
    \includegraphics[width=\columnwidth]{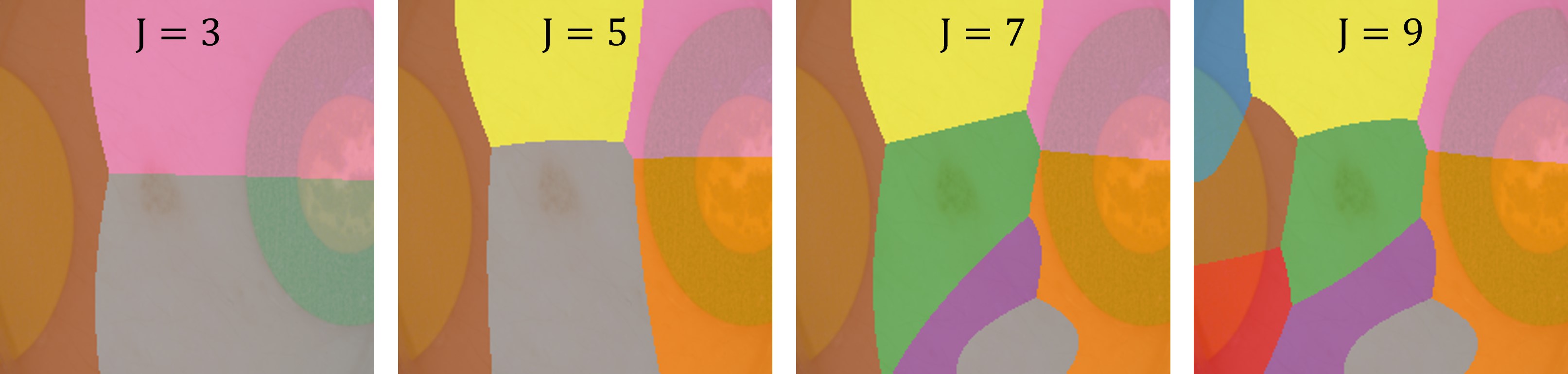}
    \caption{Illustration of clustering on the skin lesion dataset for four different choices of the number of clusters $J$. With $J = 3$, the pink and grey clusters span large regions of both the spurious bandage and the lesion. Increasing to $J = 7$ yields a more focused cluster on the true lesion area, whereas a further increase to $J = 9$ offers no substantial additional improvement.}
    \label{fig:clusters}
\end{figure}

When choosing the number of clusters for our Positive Predictive Effect Probabilistic Segmentation via Weighted Gaussian Mixtures (PPEPS-WGM), it is essential to ensure that there are theoretically enough clusters to separate spurious from core features, thereby avoiding ambiguity in the critic’s responses. While in principle there is no strict upper bound on the number of clusters, excessively increasing this number leads to the fragmentation of very small regions of core or spurious features, which in turn makes the task more difficult for the LVLM Critic \& Judge pair. The segmentation maps $C$ for varying numbers of clusters in Figure \ref{fig:clusters} illustrate that, for the skin lesion dataset and its spurious features, increasing the number of clusters beyond seven does not yield additional benefits.
In addition to the visual guidance, we conducted a short experiment to show that the verdict accuracy of the Critic \& Judge pair is quite robust to the number of clusters $J$ making it a non-critical hyperparameter. For the knee dataset, the verdict accuracy for unambiguously classifying clusters changes only marginally between five and nine clusters.
\begin{table}[h]
\centering
\caption{Comparison of the verdict accuracy for different cluster sizes $J$.}
\begin{tabular}{lcccc}
\toprule
\textbf{$J$} & \textbf{3} & \textbf{5} & \textbf{7} & \textbf{9} \\
\midrule
$\mathrm{Acc}_{\text{Verdict}}$ & 0.68 & 0.84 & 0.86 & 0.86 \\
\bottomrule
\end{tabular}
\end{table}
The verdict accuracy classifying the relevancy of a certain cluster is directly reflected in the binary correction masks (relevant yes/no) used for the alignment step and suggests in practice to pick a reasonably large number for $J$.
Following insights from previous experiments, we fixed the number of clusters to seven across all medical datasets.
\subsection{Absolute Performance on Medical Datasets} Table \ref{tab:performance} shows the absolute Average Group Accuracy (AGA) and Worst Group Accuracy (WGA) for the original model $f$ and for all shortcut mitigation approaches corresponding to the improvements visualized in Figures 9 and 10 (main paper). These results further show that the apparently larger improvement in WGA achieved by SUBG is mainly driven by a substantial reduction in overall accuracy.

\subsection{Effect of Alignment on Embedding Space}
\begin{figure}[h]
    \centering
    \includegraphics[width=0.98\columnwidth]{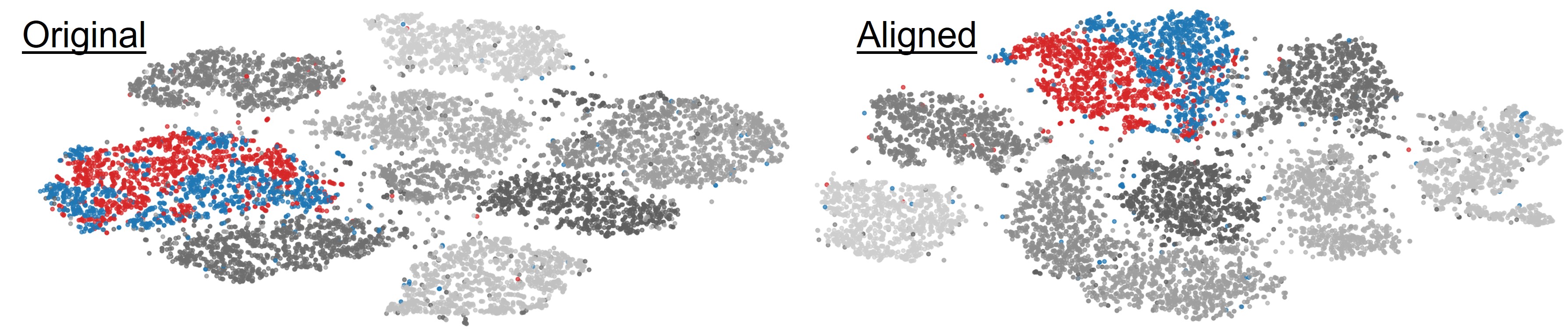}
    \caption{Test set embeddings of the DecoyMNIST MLP model before (left) and after (right) the LVLM-VA alignment step. The clusters are more separated, as the model is less affected by spurious shortcuts.}
    \label{fig:tsne_test}
\end{figure}

When observing the embeddings for the original and the aligned model on the test set (\cref{fig:tsne_test}) for DecoyMNIST, LVLM-VA reduces the model's reliance on simple shortcuts leading to better class discrimination.

\subsection{Increased Robustness to Global Perturbation}
The main scope of LVLM-VA is to reduce reliance on spurious features present during training by aligning the task model’s attribution with human expectations. Even though for global shortcuts, attribution regions may be less spatially separable and segmentation may not well isolate the spurious features, discouraging attribution to generally irrelevant regions can still improve robustness in practice.
We explicitly evaluate the effect of LVLM-VA based alignment beyond strictly localized confounders. We apply MedMNIST-C corruptions \cite{di2024medmnist} to the skin data at test time and measure the corruption-induced accuracy drop, $\Delta Acc = Acc_{\mathrm{clean}} - Acc_{\mathrm{corr}}$. We report how the alignment via LVLM-VA reduces this drop by steering the model to focus on relevant features during training, $\Delta Acc_f - \Delta Acc_{f,\mathrm{aligned}}$. Across 7 seeds, LVLM-VA significantly reduces the drop ($p < 0.05$, Wilcoxon signed-rank) for representative global corruptions that may occur in practice: GammaCorrection (+0.025), FocusBlur (+0.022), and Contrast (+0.055).

\section{Details on Experiments}
In the following, we provide additional information about the conducted experiments.

\subsection{Training Details}
Below, we describe the training settings used in our experiments, including the chosen hyperparameters and the computational environment.\\
For all experiments, we generated explanation maps $\Phi(x,y,f)$ using DeepLiftShap \cite{DeepShap}, as implemented by \citet{kokhlikyan2020captum}, with default parameters. The background dataset for DeepLiftShap consists of $25$ samples randomly drawn from the training set. Random seeds were set to $\{0,\dots, 6\}$, and the $\gamma$ parameter within the RRR loss was fixed to $0$ across all experiments. Furthermore, for all datasets, one batch comprises of $I_{x_a}=8$ alignment samples and of $I_{x_s}=64$ training samples, resulting in a total batch size of $I = 72$.

\paragraph{Computational Environment}
All experiments were implemented in Python 3.11.7 and used PyTorch 2.1.1 to train the models. We used the official OpenAI Python package (version 1.13.3) to access the GPT models. For the open-source models, we relied on inference providers available through HuggingFace. Training and fine-tuning were carried out on a machine equipped with an NVIDIA RTX~A6000 GPU and an Intel Xeon Gold 5418Y CPU with eight cores and 48GB of RAM.

\paragraph{DecoyMNIST}
For the DecoyMNIST dataset, we utilize the implementation provided by \citet{ross2017right}, which adapts the well-known MNIST dataset \cite{lecun-98}. We train the model during both the initial training phase and the alignment step for $1000$ epochs each, using a learning rate of $10^{-5}$ and the Adam optimizer \cite{kingma2014adam}. The classifier is a multilayer perceptron (MLP) with a single hidden layer of size $256$. The training set comprises $48\,000$ samples, and the test set contains $10\,000$ samples.

\paragraph{Medical Datasets}
For both medical datasets, we train a ResNet50 model \cite{he2016deep} with two output units. The original training phase and the subsequent alignment step are each run for $100$ epochs with a learning rate of $10^{-5}$ using the Adam optimizer. To determine the optimal value of $\lambda$, we perform a hyperparameter search, analogous to that used for DecoyMNIST, over a logarithmic range from $1$ to $10^{5}$. The statistical significance of improvements over the original model (before the alignment step) is assessed using a Wilcoxon signed-rank test.

\subsection{Used Prompts}
Below, we list all prompts used for the Critic \& Judge pair. For all three datasets, we provide the Critic and Judge prompts, as well as the human specifications for the respective classes. The expressions \texttt{<label>} and \texttt{<class\_description>} are instance-wise placeholders referring to the ground-truth class of the alignment sample, whereas \texttt{<cluster\_colors>} and \texttt{<num\_clusters>} denote the chosen colour set and the number of clusters, which remain fixed throughout the task.


\begin{table*}
\centering
\begin{tabular}{ccccc}
\toprule
& \multicolumn{2}{c}{\textbf{Knee Radiographs}} & \multicolumn{2}{c}{\textbf{Skin Lesions}} \\
\cmidrule(lr){2-3}\cmidrule(lr){4-5}
\textbf{Method} & \textbf{AGA} & \textbf{WGA} & \textbf{AGA} & \textbf{WGA} \\
\midrule
DFR      & $0.5714 \pm 0.0113$ & $0.0257 \pm 0.0270$ & $0.6871 \pm 0.0039$ & $0.1771 \pm 0.0482$ \\
JTT      & $0.5693 \pm 0.0271$ & $0.0471 \pm 0.0423$ & $0.7239 \pm 0.0112$ & $0.3100 \pm 0.0898$ \\
SUBG     & $0.4868 \pm 0.0148$ & $0.3457 \pm 0.0565$ & $0.6375 \pm 0.0242$ & $0.5086 \pm 0.0398$ \\
LVLM-VA  & $0.5954 \pm 0.0147$ & $0.1914 \pm 0.0631$ & $0.7493 \pm 0.0199$ & $0.4457 \pm 0.0894$ \\
original & $0.5875$ & $0.0300 $ & $0.7400$ & $0.2300$ \\
\bottomrule
\end{tabular}
\caption{Average Group Accuracy (AGA) and Worst Group Accuracy (WGA) for the original model $f$ and after the shortcut mitigation steps, for both medical datasets. The results are averaged over seven random seeds (mean ± std).}\label{tab:performance}
\end{table*}

\begin{figure*}[h]
  \centering
  \begin{tcolorbox}[
    width=.94\textwidth,
    boxrule=0.3pt,
    colback=white,
    left=6pt,right=6pt,top=6pt,bottom=6pt
  ]
  \footnotesize
  The first image is the original input image of class \texttt{<label>}, identifiable as \texttt{<class\_description>}.\\
  The second image shows \texttt{<num\_clusters>} distinct clusters \texttt{<cluster\_colors>} derived from the vision model's classification process.\\
  The third image overlays these clusters on the original image to help you locate each cluster's position.\\[2pt]

  \textbf{Analysis Instructions:}
  \begin{enumerate}\itemsep2pt
    \item First examine the original image to identify the key features of class \texttt{<label>}.
    \item For each cluster \texttt{<cluster\_colors>}:
      \begin{itemize}\itemsep2pt
        \item Describe which area of the original image it covers.
        \item Determine if this area contains relevant features for class \texttt{<label>}.
        \item Note if the cluster covers adhesive bandages (which are spurious features).
      \end{itemize}
  \end{enumerate}

  \textbf{Important Notes:}
  Adhesive bandages appear as colorful patches in the original image.\\
  These bandages are typically larger than the relevant skin lesions.\\
  Bandages are spurious features and should not be considered relevant for classification.\\[2pt]

  \textbf{For each cluster, clearly state:}
  What it covers in the original image; whether it supports correct classification (by covering relevant features); and whether it focuses on spurious features (like bandages).
  \end{tcolorbox}

  \caption{Skin Lesions - Critic.}
  \label{app:prompt-bandages}
\end{figure*}

\begin{figure*}[h]
  \centering
  \begin{tcolorbox}[
    width=.94\textwidth,
    boxrule=0.3pt,
    colback=white,
    left=6pt,right=6pt,top=6pt,bottom=6pt
  ]
  \footnotesize
  You have the task to translate text descriptions into readable JSON format for further processing.\\
  For each of the \texttt{<num\_clusters>} clusters \texttt{<cluster\_colors>}, you need to determine from the description whether it focuses on a skin lesion or not.\\[2pt]
        
  Create a JSON with "color" and "verdict" as keys where:\\
  • "color" must be one of \texttt{<cluster\_colors>}\\
  • "verdict" must be either "yes" (focuses on lesion) or "no" (does not focus on lesion)\\[2pt]
        
  \textbf{IMPORTANT:} Return only valid JSON format with an "output" key containing a list of exactly \texttt{<num\_clusters>} elements, one for each color in \texttt{<cluster\_colors>}.\\[2pt]
    
  For clarification, here are examples:\\[2pt]
    
  Example 1 Input: \\[2pt]
  \quad Analysis of Clusters:\\[2pt]

  \quad Grey:\\
  \quad - **Coverage**: Bottom center of the image.\\
  \quad - **Relevance**: Does not cover the skin lesion.\\
  \quad - **Spurious Features**: Covers part of a colored patch (bandage).\\
  \quad - **Conclusion**: Not relevant.\\[2pt]
  \dots

  Example 1 JSON:\\
  \{\{\\
  \quad "output": [\\
  \quad\ \ \{\{\\
  \quad\ \ \ \ "color": "grey",\\
  \quad\ \ \ \ "verdict": "no"\\
  \quad\ \ \}\}\\
  \dots]\\
  \}\}\\
  \end{tcolorbox}

  \caption{Skin Lesions - Judge}
  \label{app:json-prompt-skin}
\end{figure*}

\begin{table*}[h]
  \centering
  \caption{Skin Lesions - Human Specifications per Class}
  \label{tab:skin-lesion-class-descriptions}
  \begin{tabularx}{0.95\textwidth}{@{}>{\raggedright\arraybackslash}p{3.0cm}X@{}}
    \toprule
    \textbf{Class} & \textbf{Description} \\
    \midrule
    \texttt{benign} &
    Non-cancerous skin growths that typically present with symmetrical shapes, clear borders, and uniform brown or black coloring with a small diameter, i.e., covering only a small part of the image. \\
    \texttt{malignant} &
    Cancerous skin lesions characterized by asymmetry, irregular borders, brown or black color variations with multiple distinct colors, and diameters typically covering a large area of the image. \\
    \bottomrule
  \end{tabularx}
\end{table*}

\begin{figure*}[h]
  \centering
  \begin{tcolorbox}[
    width=.94\textwidth,
    boxrule=0.3pt,
    colback=white,
    left=6pt,right=6pt,top=6pt,bottom=6pt
  ]
  \footnotesize
  The first image is the original input image of class \texttt{<label>}, identifiable as \texttt{<class\_description>}.\\
  The second image shows \texttt{<num\_clusters>} distinct clusters \texttt{<cluster\_colors>} derived from the vision model's classification process.\\
  The third image overlays these clusters on the original image to help you locate each cluster's position.\\[2pt]

  \textbf{Analysis Instructions:}
  \begin{enumerate}\itemsep2pt
    \item First examine the original image to identify the key features of class \texttt{<label>}.
    \item For each cluster \texttt{<cluster\_colors>}:
      \begin{itemize}\itemsep2pt
        \item Describe which area of the original image it covers.
        \item Determine if this area contains relevant features for class \texttt{<label>}.
        \item Note if the cluster covers radiographic markers (letters \texttt{R} or \texttt{L}), which are spurious features.
      \end{itemize}
  \end{enumerate}

  \textbf{Important Notes:}
  Radiographic markers appear as letters \texttt{R} or \texttt{L} in the original image, indicating right or left leg.\\
  These markers are typically located at the edges of the image, outside the joint area.\\
  Markers are spurious features and should not be considered relevant for classification.\\[2pt]

  \textbf{For each cluster, clearly state:}
  What it covers in the original image; whether it supports correct classification (by covering relevant features); and whether it focuses on spurious features (like radiographic markers).\\[4pt]

  \end{tcolorbox}

  \caption{Knee Radiographs - Critic}
  \label{app:prompt-radiographic}
\end{figure*}

\begin{figure*}[h]
  \centering
  \begin{tcolorbox}[
    width=.94\textwidth,
    boxrule=0.3pt,
    colback=white,
    left=6pt,right=6pt,top=6pt,bottom=6pt
  ]
  \footnotesize
  You have the task to translate text descriptions into readable JSON format for further processing.\\
  For each of the \texttt{<num\_clusters>} clusters \texttt{<cluster\_colors>}, you need to determine from the description whether it is relevant and does not focus on spurious features.\\
  If cluster includes both a part of relevant features and a spurious feature, it should be assigned "no" (not relevant), unless the covered part of the relevant features is large, then it should be assigned "yes".\\[2pt]

  Create a JSON with "color" and "verdict" as keys where:\\
  • "color" must be one of \texttt{<cluster\_colors>}\\
  • "verdict" must be either "yes" (might focus on relevant areas of the knee) or "no" (focuses on spurious features)\\[2pt]
        
  \textbf{IMPORTANT:} Return only valid JSON format with an "output" key containing a list of exactly \texttt{<num\_clusters>} elements, one for each color in \texttt{<cluster\_colors>}.\\[2pt]
    
  For clarification, here are examples:\\[2pt]
    
  Example 1 Input: \\[2pt]
  \quad Analysis of Clusters\\

  \quad Grey Cluster\\
  \quad - **Coverage**: Small part of the knee.\\
  \quad - **Relevance**: Might be relevant as it does not cover a spurious feature.\\
  \quad - **Spurious Features**: Does not cover any spurious features.\\[2pt]
  \dots

  Example 1 JSON:\\
  \{\{\\
  \quad "output": [\\
  \quad\ \ \{\{\\
  \quad\ \ \ \ "color": "grey",\\
  \quad\ \ \ \ "verdict": "yes"\\
  \quad\ \ \}\}\\
  \dots]\\
  \}\}\\
  \end{tcolorbox}

  \caption{Knee Radiographs - Judge}
  \label{app:json-prompt-knee}
\end{figure*}

\begin{table*}[h]
  \centering
  \caption{Knee Radiographs - Human Specifications per Class}
  \label{tab:oa-class-descriptions}
  \begin{tabularx}{0.95\textwidth}{@{}>{\raggedright\arraybackslash}p{3.2cm}X@{}}
    \toprule
    \textbf{Class} & \textbf{Description} \\
    \midrule
    \texttt{no osteoarthritis} &
    Healthy knee joints without osteoarthritis, typically characterized by normal joint spacing, absence of bone deformities, smooth joint surfaces, and clear cartilage without visible deterioration. \\
    \texttt{osteoarthritis} &
    Knee joints affected by osteoarthritis, characterized by narrowed joint spaces, bone spurs (osteophytes), noticeable cartilage degradation, sclerosis (increased bone density), and visible joint irregularities or deformities. \\
    \bottomrule
  \end{tabularx}
\end{table*}

\begin{figure*}
  \centering
  \begin{tcolorbox}[
    width=.94\textwidth,
    boxrule=0.3pt,
    colback=white,
    left=6pt,right=6pt,top=6pt,bottom=6pt
  ]
  \footnotesize

  The following images include hand written digits.\\
  The first image is the original input image of class \texttt{<label>}, which can be recognized as \texttt{<class\_description>}.\\
  The second image is a visualization map indicating different clusters considered important for classifying class \texttt{<label>}.\\
  The third image is a visualization map from class \texttt{<label>} overlaid in the original image to support you in relating locations between both images.\\
  In some of the images spurious decoys in the corner are introduced to confuse the model generating the clustered visualization maps and the squares in the corner do not represent one of the classes and should be considered spurious and not be considered within the visualization maps.\\
  The visualization map consist of \texttt{<num\_clusters>} clusters with the colors \texttt{<cluster\_colors>}, where each cluster describes an area of focus from the original image.\\
  First, examine the original image to identify which parts belong to class \texttt{<label>}.\\
  Then, look at the second image to see the \texttt{<num\_clusters>} clusters for class \texttt{<label>}.\\
  For each cluster \texttt{<cluster\_colors>}, describe the area where the model focuses to predict class \texttt{<label>}.\\
  Determine whether each cluster is within the boundaries of the class \texttt{<label>} using the third image.\\
  A cluster supports the correct prediction only if it fully or partially focuses on an area within the class \texttt{<label>}.\\
  If a cluster is outside the class structure, clearly state that this cluster does not support the correct prediction.\\
  Do not provide introductory sentences.\\
  Consider the following three examples: \dots
  \end{tcolorbox}

  \caption{DecoyMNIST - Critic}
  \label{app:prompt}
\end{figure*}

\begin{figure*}[h]
  \centering
  \begin{tcolorbox}[
    width=.94\textwidth,
    boxrule=0.3pt,
    colback=white,
    left=6pt,right=6pt,top=6pt,bottom=6pt
  ]
  \footnotesize
   You have the task to translate the responses of a large vision language model (LVLM) into readable JSON format for further processing.\\
  The task of the LVLM was, for each of the \texttt{<num\_clusters>} clusters in the second and third image, to identify whether the focus aligns with any part of the digit depicted in a first image or not.\\
  First, you read through the LVLM response. Then you identify for each of the clusters \texttt{<cluster\_colors>} whether the focus was on the digit or not.\\
  Then you construct a valid JSON with \texttt{"color"} and \texttt{"verdict"} as keys. The \texttt{"color"} key should strictly be one of \texttt{<cluster\_colors>}. The \texttt{"verdict"} key should strictly be either \texttt{"yes"} or \texttt{"no"}.\\
  For example, \texttt{"color": "red"} and \texttt{"verdict": "yes"} means that the red cluster did focus on the digit. \texttt{"color": "blue"} and \texttt{"verdict": "no"} means that the blue cluster did not focus on the digit.\\
  \textbf{IMPORTANT:} Please make sure to only return in valid JSON format, with the \texttt{"output"} key as a list of JSON. The list should strictly contain \texttt{<num\_clusters>} elements, one for every cluster in \texttt{<cluster\_colors>}.\\
  For clarification, here are a few examples: \\
    \textbf{Examples}\\[-2pt]
Example 1 Input:
  Pink: The cluster covers the vertical line of the digit 7 and no decoy
        square in the corner. It is relevant.
  Brown: The cluster fully covers a decoy square in the lower right corner.
         It is not relevant.
  Yellow: The cluster covers the horizontal line of the digit 7 and no decoy
          square in the corner. It is relevant.
Example 1 JSON:
{
  "output": [
    { "color": "pink",   "verdict": "yes" },
    { "color": "brown",  "verdict": "no"  },
    { "color": "yellow", "verdict": "yes" }
  ]
}
\dots
===== END OF EXAMPLES =====
  \end{tcolorbox}

  \caption{DecoyMNIST - Judge}
  \label{app:json-prompt}
\end{figure*}

\begin{table*}[h]
  \centering
  \caption{DecoyMNIST - Human Specifications per Class}
  \label{tab:class-descriptions}
  \begin{tabularx}{0.95\textwidth}{@{}>{\raggedright\arraybackslash}p{1.2cm}X@{}}
    \toprule
    \textbf{Class} & \textbf{Description} \\
    \midrule
    0 & A closed, continuous loop with no starting or ending point, representing a circle or oval shape. \\
    1 & A single, straight vertical line, typically with a small base or serif at the bottom. \\
    2 & A curved line starting from the top, forming an open loop to the right, and then descending in a diagonal line toward the left. \\
    3 & Two small, open, curved loops stacked vertically, each curving to the right, connected in the middle. \\
    4 & A vertical line with an angled horizontal line starting from its midpoint, and a diagonal line connecting the top of the vertical line to the bottom of the horizontal line. \\
    5 & A horizontal line at the top connected to a vertical line descending downward, which then curves sharply to the left and forms an open loop. \\
    6 & A vertical line starting from the top, curving downward to the left, and forming a closed loop at the bottom. \\
    7 & A horizontal line at the top connected to a diagonal line that descends toward the left, with no curves or loops. \\
    8 & Two distinct loops one on the top and one on the bottom connected in the middle. \\
    9 & A small loop at the top with a vertical line descending downward from the loop's right side. \\
    \bottomrule
  \end{tabularx}
\end{table*}

\end{document}